\documentclass[final]{siamltex}
\usepackage{algorithm,algpseudocode}
\usepackage{latexsym, graphicx, epsfig, amsmath, amsfonts,amssymb,bm}
\usepackage{caption, subcaption}
\usepackage{epstopdf}
\usepackage{booktabs}
\usepackage{array}
\usepackage{color}
\usepackage{lineno}
\usepackage{url}

\usepackage{graphicx}
\usepackage{grffile}
\usepackage{subcaption}

\def\be{\begin{equation}}
\def\ee{\end{equation}}

\def\x{\mathbf{x}}

\def\w{\mathbf{w}}
\def\f{\mathbf{f}}
\def\z{\mathbf{z}}
\def\zh{\hat{\z}}
\def\p{\mathbf{p}}
\def\q{\mathbf{q}}
\def\I{\mathbf{I}}

\def\xt{\widetilde{\x}}
\def\X{\mathbf{X}}

\def\Y{\mathbf{Y}}
\def\Rb{\mathbf{R}}
\def\Rs{\mathbb{R}}

\def\M{\mathcal{M}}
\def\Mb{\mathbf{M}}
\def\A{\mathbf{A}}

\def\K{\mathbf{K}}
\def\F{\mathbf{F}}

\def\N{\mathbf{N}}
\def\Pi{\mathbf{\Phi}}

\def\a{\boldsymbol{\alpha}}

\def\ea{e_{1p}}
\def\eb{e_{2p}}
\def\ec{e_{3p}}

\newcommand{\argmin}{\operatornamewithlimits{argmin}}

\title{Modeling Unknown Dynamical Systems with Hidden Parameters}

\author{Xiaohan Fu\footnotemark[1]\and WeiZe Mao\footnotemark[2]\and Lo-Bin Chang\thanks{Department of Statistics,
    The Ohio State University, Columbus, OH 43210, USA. {\tt
      Emails: fu.688@osu.edu, lobinchang@stat.osu.edu.}} \and Dongbin
       Xiu\thanks{Department of Mathematics,
		The Ohio State University, Columbus, OH 43210, USA.
		{\tt Email: xiu.16@osu.edu.}
		Funding: This work was partially supported by AFOSR FA9550-18-1-0102.}
}

\begin{document}
\maketitle
\begin{abstract}
We present a data-driven numerical approach for modeling unknown dynamical systems with missing/hidden
parameters. The method is based on training a deep neural network (DNN) model for the unknown system using its trajectory 
data. A key feature is that the unknown dynamical system contains system parameters that are completely hidden, in the sense that
no information about the parameters is available through either the measurement trajectory data or our prior
knowledge of the system. We demonstrate that
by training a DNN using the trajectory data with sufficient time history, the resulting DNN model can accurately model the unknown dynamical system.
For new initial conditions associated with new, and unknown, system parameters, the DNN model can produce accurate system predictions
over longer time. 
\end{abstract}

\section{Introduction} \label{sec:intro}

There has been a growing interest in learning unknown dynamical systems using observational data. A common approach is to construct a mapping from the state variables to their time derivatives. Various numerical approximation techniques can be used to construct such a mapping. These include sparse regression, polynomial approximations, model  selection, Gaussian process regression (\cite{Mangan20170009,raissi2017machine,brunton2016discovering,schaeffer2017extracting, rudy2017data,WuXiu_JCPEQ18, WuQinXiu2019}), to name a few. More recently, deep neural networks (DNNs) are adopted to construct the mapping. Studies have empirically demonstrated the ability of DNN to model ordinary differential equations (ODEs)  \cite{raissi2018multistep,qin2018data,rudy2018deep} and partial differential equations (PDEs) \cite{long2017pde,raissi2017physics1,raissi2017physics2,raissi2018deep,long2018pde,sun2019neupde}. A notable recent development is to model the mapping between two system states separated by a short time (\cite{qin2018data}). This approach essentially models the underlying flow map of the unknown system, and is notably different from the earlier approach of modeling the map between the state variables and their time derivatives. 
The flow map based approach eliminates the need for temporal derivative data, which are often difficult to acquire in practice and subject to larger errors. Once an accurate DNN model for the flow map is constructed, it can be used as an evolution operator to conduct
system predictions. In particular, residual network (ResNet), developed in image analysis community (\cite{he2016deep}), was found to be suitable for recovering the flow map (\cite{qin2018data, ChenXiu_JCP20}). Since its introduction
(\cite{qin2018data}), the flow map based DNN modeling approach has been extended to modeling of non-autonomous
dynamical systems (\cite{QinCJD_Nonauto21}), parametric dynamical systems (\cite{QinCJX_IJUQ20}), partially observed dynamical systems (\cite{fu2020}), as well as partial differential equation (\cite{WuXiu_JCP20}). 


The focus of this paper is on a different type of data driven modeling problems. We assume that the target unknown
dynamical system is parameterized by a set of parameters that are completely hidden, in the sense that no prior
knowledge about the form, or even the existence, of the parameters is available. The only available information of the dynamical system
is in the form of trajectory
data of its state variables. The trajectory data are also parameterized, in an unknown manner, by the hidden parameters.
Our goal is to construct a predictive model
of the underlying dynamical system by using only the trajectory data. Once the predictive model is constructed, it shall
be able to produce accurate predictions of the system states over time, for any given initial conditions that are parameterized by the hidden parameters in unknown manner. The distinct feature of this work is that no knowledge of the system parameters is assumed to be available,
not in the (unknown) governing equations or in the trajectory data (for training or prediction). This is often the case for many complex systems,
whose dynamics are controlled by a large, and sometimes unknown, number of parameters that are not measurable.

The method proposed in this paper is motivated by the work of \cite{fu2020}, which studied modeling of partially observed
dynamical systems where trajectory data of only a subset of the state variables are available.
While the celebrated Mori-Zwanzig (MZ) formulation (\cite{mori1965, zwanzig1973}) defines a closed-form dynamical system for the observed state variables, the MZ system is intractable for practical computations as it involves a memory integral of an unknown kernel function.
Upon assuming a finite effective memory length, a DNN structure with explicit incorporation of ``past memory'' was proposed in \cite{fu2020} and shown to be highly effective for learning and modeling partially
observed systems. Compared to other DNN strutures with memory gates, e.g. LSTM, the DNN structure from \cite{fu2020} is notably simpler and serves as a direct approximation of the Mori-Zwanzig formulation.

In this paper, we adopt the DNN structure developed in \cite{fu2020} and demonstrate that it can be used to model unknown dynamical systems with hidden parameters. The theoretical motivation is that the hidden parameters can be viewed as a
set of unobserved state variables with trivial dynamics. Consequently the DNN structure from \cite{fu2020} becomes applicable. Moreover, for long-term prediction accuracy and stability, we introduce a recurrent structure during network training. Once the DNN model is constructed, it is able to produce accurate system predictions over longer time, for any given
initial conditions containing unknown hidden parameters.


\section{Setup and Preliminaries} \label{sec:setup}

Let us consider a dynamical system
\be \label{govern}
\frac{d\xt}{dt}(t; \a)=\f(\xt, \a),\qquad \xt(0; \a) = \xt_0,
\ee
where $\xt \in\Rs^n$ are state variables and $\a\in\Rs^d$ are 
system parameters.  We assume that the form of
the governing equations, which manisfests itself via $\f:\Rs^n\times
\Rs^d\to\Rs^n$, is unknown. 
More importantly, we assume that the information about the system parameters
$\a$ is not available. In fact, even the dimensionality $d$ of $\a$
can be unknown.

\subsection{Learning Objective}

We assume trajectory data are available for the state variables $\xt$.
Let $N_T$ be the total number of observed trajectories. For
each $i$-th trajectory, we have
\be \label{X}
\X^{(i)} = \left\{ \xt\left(t^{(i)}_k\right)\right\}, \qquad k=1,\dots, K^{(i)},
 \quad i=1,\dots, N_T,
\ee
where $\{t_k^{(i)}\}$ are discrete time instances at which the data are
available, and $K^{(i)}$ is the total number of data entries in the $i$-th
  trajectory. Note that each $i$-th trajectory is associated with an
  initial condition $\xt_0^{(i)}$ and system parameters $\a^{(i)}$,
  both of which are unknown.

Our goal is to construct an accurate numerical model,
$\M$ for the system \eqref{govern}, by using the data set \eqref{X}.
More specifically, let
$$
0 = t_0< \cdots < t_N = T
$$
be a sequence of time instances with a finite horizon $T>0$. This
will be our prediction time stencil. We seek a predictive model
$\M$ such that, for any given initial condition $\x_0$, which is
associated with
an unknown system parameter $\a$, 
the model prediction is an accurate approximation of the true system,
in the sense that
\be
\M(t_k; \x_0, \a) \approx \xt(t_k; \x_0, \a), \qquad
k=1,\dots, N,
\ee
with satisfactory accuracy.

\subsection{Related Study}

Our topic is related to, and extends, two recent studies on modeling
dynamical systems. The first related study is on recovering unknown
deterministic dynamical systems.
 When data of the state variables $\x$ are available, it was shown in
 \cite{qin2018data} that residual network (ResNet) can be
used to construct a predictive model. In fact, for autonomous systems,
the ResNet based DNN model is an exact
integrator of the underlying system. It is a one-step predictive model
and consequently requires only trajectory data of two consecutive data
entries.
For parameterized systems, when the parameter $\a$ are known from the
trajectory data, the ResNet model can be modified to incorporate more
input neurons to represent the system parameters $\a$. See
\cite{QinCJX_IJUQ20} for detail.

Another related study is on modeling unknown dynamical systems with
partially observed state variables.
Let $\x^\top = (\z^\top, \w^\top)$ be the full set of state variables,
where $\z \in \Rs^n$ is the subset of the state variables with
available data, and $\w \in \Rs^d$ is the subset of  missing
variables. Based on the celebrated Mori-Zwanzig (MZ) formulation
(\cite{mori1965},\cite{zwanzig1973}), the evolution of $\z$ follows a
generalized Langevin equation,
\be \label{eq:MZ}
\frac{d}{dt}\z (t)=\Rb(\z(t)) + \int_0^t\K(\z(t-s),s)ds + \F(t, \x_0),
\ee
which involves a Markovian term $\Rb$, a memory integral with kernel $\K$
and a random term $\F$ involving the unknown initial condition.
Upon making an assumption on finite effective memory,
a discrete approximate Mori-Zwanzig equation was proposed in \cite{fu2020},
\be \label{DAMZ}
\left.\frac{d}{dt}{\zh} (t)\right|_{t=t_n} = \left.\Rb({\zh}(t))\right|_{t=t_n} + \Mb(\zh_{n-n_M}, \dots, \zh_{n-1}, \zh_n),
\ee
where $\zh_n=\zh(t_n)$ is the solution at time $t_n = n\Delta$ over a
constant time step $\Delta$, $n_M$ is the number of memory terms. A
DNN structure to explicitly account for the
memory terms was then proposed in  \cite{fu2020} and shown to be
highly effective and accurate. 


\section{Method Description} \label{sec:method}

In this section, we describe the detail of our proposed deep learning
approach for system with hidden parameters. The distinct feature of
our work is that not only are the system equations unknown, the
associated system parameters remain completely unknown throughout the
modeling and prediction process.

\subsection{Motivation} \label{sec:motivation}

For the unkonwn system with missing/hidden parameters \eqref{govern},
one can view it in an alternative form,
\be \label{govern2}
\left\{
\begin{split}
  &\frac{d\xt}{dt}=\f(\xt, \a),\qquad \xt(0) = \xt_0, \\
  &\frac{d\a}{dt} = \mathbf{0}, \qquad \a(0) = \a.
\end{split}
\right.
\ee
If one treats $\a$ also as state variables with trivial dynamics and
views  $\widetilde{\X} = (\xt^\top, \a^\top)^\top$ as the complete set of state
variables, the data set \eqref{X} on $\xt$ then represents the data of a
subset of the full variable set $\widetilde{\X}$. From this
perspective, the memory based DNN structure, designed in \cite{fu2020}
for partially observed systems, becomes applicable. Hereafter we will
employ the DNN structure of  \cite{fu2020} and modify it to suit our modeling needs.

\subsection{Network Structure} \label{sec:NN}

Our basic DNN structure consists of a forward block and a recurrent
block. For notational convenience, hereafter we shall assume a constance time step
\be \label{Delta}
\Delta \equiv t_{k+1}^{(i)} - t_{k}^{(i)}, \qquad \forall k=1,\dots,
K^{(i)}-1, \quad i=1,\dots, N_T,
\ee
for all the trajectory data, as well as for the 
prediction time stencil. (Variable time steps can be
readily incorporated into the DNN model as an additional input.
See \cite{QinCJX_IJUQ20} for detail.)

\subsubsection{Forward Block} \label{sec:block}

\begin{figure}
	\centering
	\includegraphics[scale=0.4]{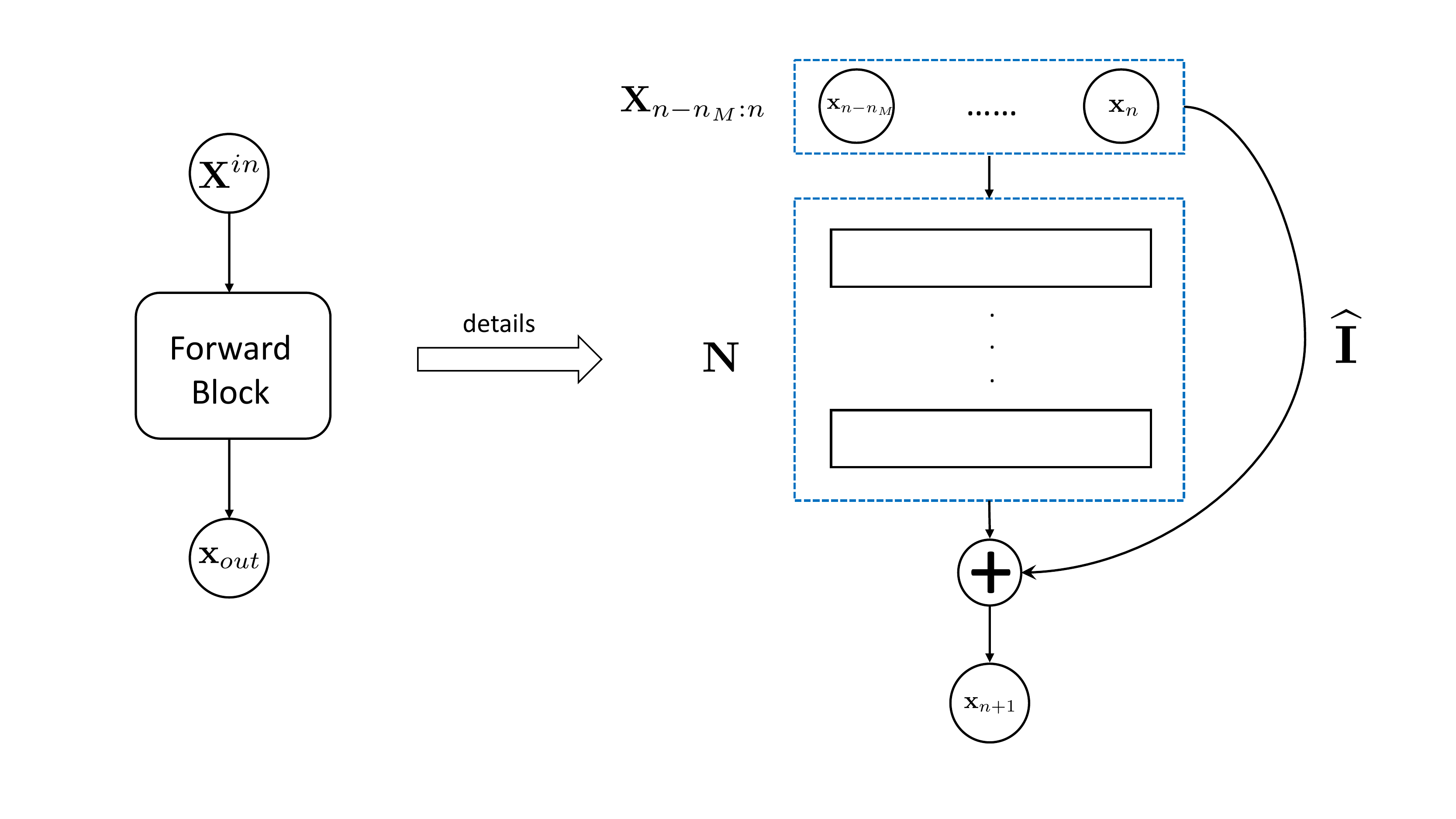}
	\caption{Illustration of forward block.}
	\label{fig:Net}
      \end{figure}

The forward block of our DNN model is similar to the DNN with
memory model developed in \cite{fu2020}. 
The structure of forward block is illustrated in Figure
\ref{fig:Net}. Its input layer incorporates $(n_M+1)$ state vectors $\x$,
each of which has size $n$. The output layer incorporates a single
state vector $\x$ of length $n$. A standard fully connected
feedforward network (FFN) serves as the mapping from the input layer
to the output layer. We use $\N$ to denote the mapping operator
defined by the FNN. An operator $\widehat{\mathbf{I}}$ is introduced
to the input layer and then applied to the output of the FNN. This is
to achieve the ResNet-like operation. 
      
More specifically, the dimension of the input layer, i.e., the number
of neurons in the input layer,  is
\be \label{D}
D= n\times (n_M+1).
\ee
We write 
\be \label{ZZ}
\X =\left(\x_n^\top, \x_{n-1}^\top, \dots, \x_{n-n_M}^\top\right)^\top \in \Rs^{D}
\ee
as the input vector to the DNN model.  We then define
$\widehat{\mathbf{I}}$ as 
a $(n\times D)$ matrix, 
$$
\widehat{\mathbf{I}} = \left[\mathbf{I}_n, \mathbf{0}, \dots,
  \mathbf{0}\right],
$$
    where the size $(n\times n)$ identity matrix $\mathbf{I}_n$ is
    concatenated by $n_M$ zero matrices of size $(d\times d)$. 
The fully connected FNN connecting the input and output layers then
defines a mapping operator
    \be \label{Noper}
    \N(\cdot; \Theta):\Rs^D\to \Rs^n,
    \ee
    where $\Theta$ is the hyperparameter set associated with the FNN.
Upon applying the operator  $\widehat{\mathbf{I}}$ to the input and
re-introducing it at the output of the FNN operation, our DNN model then
defines the following operation
    \begin{equation} \label{Net}
      \x^{out} =  \left[\widehat{\mathbf{I}} +
        {\N}\right]\left(\X^{in}\right),
    \end{equation}
    which in turn can be written as
\be \label{Net_scheme}
\x_{n+1} = \x_{n} + \N(\x_n, \x_{n-1}, \dots, \x_{n-n_M}; \Theta), \qquad n\geq n_M.
\ee

We remark that $n_M\geq 0$ is the number of
memory steps included in our DNN model. Let $T_M = n_M\times
\Delta$. This shall be the length of the effective memory, a concept
introduced in \cite{fu2020}. The choice of $T_M$ is problem dependent
and requires certain prior knowledge/experience about the underlying
system. Sometimes trial-and-error is also necessary. Such practice is
not uncommon in
many aspects of numerical analysis, for example, choices of domain size
and grid size. Note that $n_M=0$ represents the memory-less case,
which reduces the DNN back to the standard ResNet structure used for
modeling complete system \cite{qin2018data}.

\subsubsection{Recurrent Block} \label{sec:multiblock}

The forward DNN block discussed in the previous section is essentially the same DNN structure developed
in \cite{fu2020}, for modeling systems with missing
variables. In principle, it is also applicable for modeling
systems with hidden parameters, as motivated in Section
\ref{sec:motivation}. However, during our initial numerical
experimentations, we have repeatedly discovered 
that it lacks sufficient long-term numerical stability.
To mitigate the numerical instability, we thus introduce a recurrent structure, in
conjunction with the forward block, in our final DNN model.

\begin{figure}
	\centering
	\includegraphics[width=\textwidth]{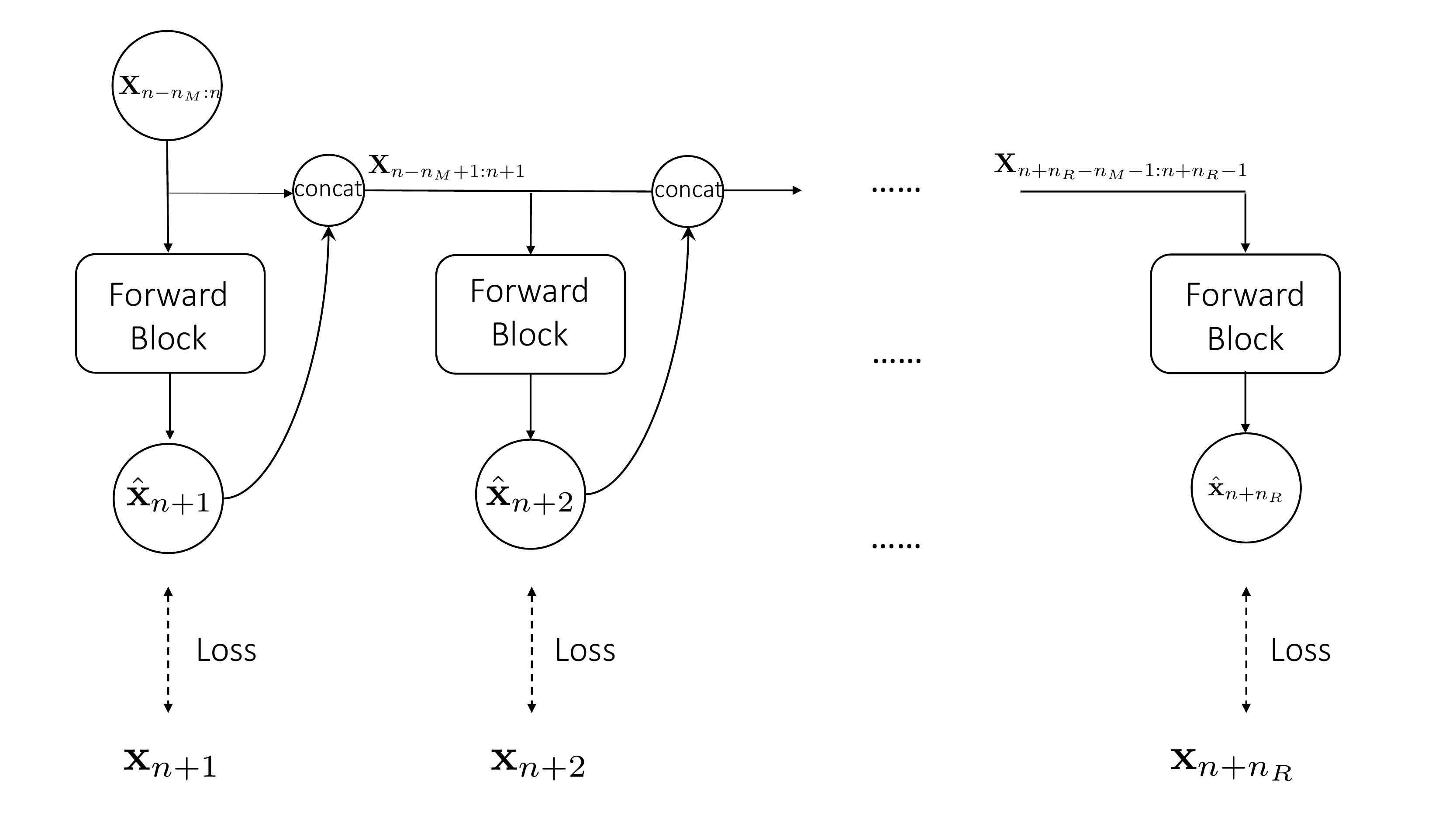}
	\caption{Illustration of recurrent-forward-block structure with $n_R$ recurrent steps.}
	\label{fig:multiblock}
\end{figure}

The structure of the recurrent block is illustrated in Figure
\ref{fig:multiblock}, where $n_R\geq 1$ is the number of recurrent steps. The trivial
case of $n_R=1$ reduces the DNN model to the forward block structure in
the previous section. The recurrent blocks are to
recursively apply the forward DNN block over $n_R$ time steps and compute the loss
function using the outputs of the $n_R$ steps.

Let
\be
\X_{i:j} = \left(\x_{i}^\top, \dots,
  \x_{j}^\top\right)^\top, \qquad j\geq i,
\ee
be the concatenated state variable vectors from $\x_i$ to $\x_j$, with
$j\geq i$.
Our final DNN model with $n_R$ recurrent step can then be defined as,
for any time $t_n$ with $n\geq n_M$,
\be \label{multischeme}
\begin{cases} 
	\X^{in} = \X_{n-n_M:n}, \\
	\x_{k+1} = \left[\widehat{\mathbf{I}} + \N\right]\left(\X_{k-n_M:k}\right), \qquad k=n,\dots, n+n_R,\\
	\X^{out} = \X_{n+1:n+n_R}.
\end{cases}
\ee
Note that the $n_R$ forward blocks share the same parameter set
$\Theta$. In other words, it is the same forward block that is applied
recurrently $n_R$ times.
The input of the entire DNN network is the same as that of the
non-recurrent forward block, $\X_{n-n_M:n} = \left(\x_{n-n_M}^\top,
  \dots, \x_{n}^\top\right)^\top$, $n_M+1$ steps of solution vectors. The output of the DNN is a sequence
of $n_R$ steps of the outputs of the forward block.

\subsection{Network Training and
  Predictive Modeling} \label{sec:training}

The DNN model \eqref{multischeme} effectively defines a mapping
\be
\X^{out} = \mathcal{N}(\X^{in}; \Theta),
\ee
where $\X^{in}$ consists of $n_M+1$ steps of the state variables $\x$, and
$\X^{out}$ consists of $n_R$ steps of the state variables. Therefore, to
train the DNN model, we require state variable trajectories of length
at least $n_{tot} = n_M+n_R+1$.

Let us assume that each $i$-th trajectory in our data set
\eqref{X} has its number of entries satisfying $K^{(i)}\geq
n_{tot}$ entries. (In other words, the 
trajectories with less number of entries are already eliminated from the data set.)
We then randomly select a piece of $n_{tot}$ number of consecutive entries from
the trajectory and re-group them into two segments: the
first $n_M+1$ entries {\em vs.} the last $n_R$ entries:
\be \label{group}
\left\{ \X^{(i)}, \Y^{(i)}\right\},
\quad
\ee
where
\be
\begin{split}
	\X^{(i)} &= \left[\x\left(t^{(i)}_{k}\right)^\top, \dots,
	\x\left(t^{(i)}_{k+n_M}\right)^\top\right]^\top  \\
	\Y^{(i)} &= \left[\x\left(t^{(i)}_{k+n_M+1}\right)^\top, \dots,
	\x\left(t^{(i)}_{k+n_M+n_R}\right)^\top\right]^\top.
\end{split}
\ee

This random selection
procedure is repeated for all the $N_T$ trajectories in the data set
\eqref{X}. Note that for each $i=1,\dots, N_T$ trajectory, it is
possible to select more than one such groupings whenever $K^{(i)}> n_{tot}$.
Upon conducting the random sequence selection for all the trajectories
in \eqref{X}, we obtain a collection of the grouping
\eqref{group}. After re-ordering all the selected groupings with a
single index, we obtain the training data set for our DNN model,
\be \label{data_set}
\mathcal{X} = \left\{ \X_j, \Y_j\right\}, \qquad j=1,\dots, J,
 \quad
 \ee
 where $J$ is the total number of the data
 groupings.
 (Note that at this stage the information of the $i$-th
 trajectory, from which the grouping $\left\{ \X_j, \Y_j\right\}$
 is originated,  is not important.)

 Our DNN model training is then conducted by minimizing the following
 mean squared loss
     \begin{equation} \label{eq:loss}
      \Theta^* =\argmin_\Theta\frac{1}{J}\sum_{j=1}^J\left\|
        \mathcal{N}(\X_j^{in}; \Theta)-\Y_j\right\|^2.
    \end{equation}
   %
    Upon finding the optimal network parameter $\Theta^*$, we obtain
   our trained network model in the form of \eqref{multischeme},
    \begin{equation} \label{TheNet}
     \x^{out} = \left[\widehat{\mathbf{I}} +
       {\N}(\cdot; \Theta^*)\right]\left(\X^{in}\right),
   \end{equation}
   where the optimized parameter $\Theta^*$ will be omitted hereafter,
   unless confusion arises otherwise.
   
 The trained DNN model defines a predictive model for the unknown dynamical
 system \eqref{govern} with hidden parameters. It requires $n_M+1$
 initial conditions. Once given a sequence of $n_M+1$ state variables $\x$,
 which are associated with unknown parameters $\a$, the DNN model is
 able to conduct one-step prediction iteratively for the system state,
 corresponding to the same (and yet still unknown) parameters
 $\a$. More specifically, the predictive scheme takes the following
 form: for any unknown hidden parameter $\a$,
 \be \label{Model}
\left\{
\begin{split}
& \x_k = \x(t_k; \a), \qquad k=0,\dots, n_M, \\
  &\x_{n+1} = \x_{n} + \N(\x_n, \x_{n-1}, \dots, \x_{n-n_M}; \Theta^*), \qquad n\geq n_M.
\end{split}
\right.
\ee

\section{Numerical Examples} \label{sec:examples}

In this section, we present four numerical examples to examine the
performance of the proposed method.
The examples include (1) a nonlinear pendulum system with 2 hidden
parameters, (2) a larger linear system with 100 hidden parameters; (3)
a nonlinear chemical reactor system with one hidden parameter that
induces bifurcation in the system behavior; and (4) a nonlinear system
for modeling cell signaling cascade with 12 hidden parameters.
In all the examples, the underlying ``true'' models are known and 
used only to generate the training data sets. Note that in the training
data sets, only the solution trajectories are recorded; the corresponding
parameter values are not
recorded. By doing so, the parameters in the true models remain
completely hidden from the DNNs.
To validate the trained DNN predictive models, we use the corresponding
true models to generate a set of initial conditions that are not in the
training data sets and with the associated parameter values
hidden. The DNN predictive models are then used to produce system
predictions over longer time horizon and compared against the
reference solutions generated by the true models.

In all the examples here, the time step is fixed at $\Delta =
0.02$. The number of memory steps $n_M$ and recurrent steps $n_R$ are
problem dependent and determined numerically by gradually increasing
the values till converged numerical results are obtained.
Unless otherwise noted, the DNNs used in the examples
consist of 3 hidden layers, each of which with 30 neurons, and have
rectified linear unit (ReLU) activation function.

\subsection{Example 1: Nonlinear Pendulum System}
We first consider a small nonlinear system, the damped pendulum system, 
\begin{equation}
	\begin{cases}
		\dot x_1 = x_2,\\
		\dot x_2 = -\alpha x_2 - \beta \sin x_1,
	\end{cases}       
      \end{equation}
      where the system parameters $\a=(\alpha,\beta)^\top$ are treated
      as hidden and confined to a region $D_{\a} = [0.05,0.15] \times
      [8,10]$. The domain of interest for the state variables is set
      as $D_{\x} = [-0.5,0.5] \times [-1.6,1.6]$.

\begin{figure}
	\centering
	\begin{subfigure}[b]{0.5\textwidth}
		\includegraphics[width=\textwidth]{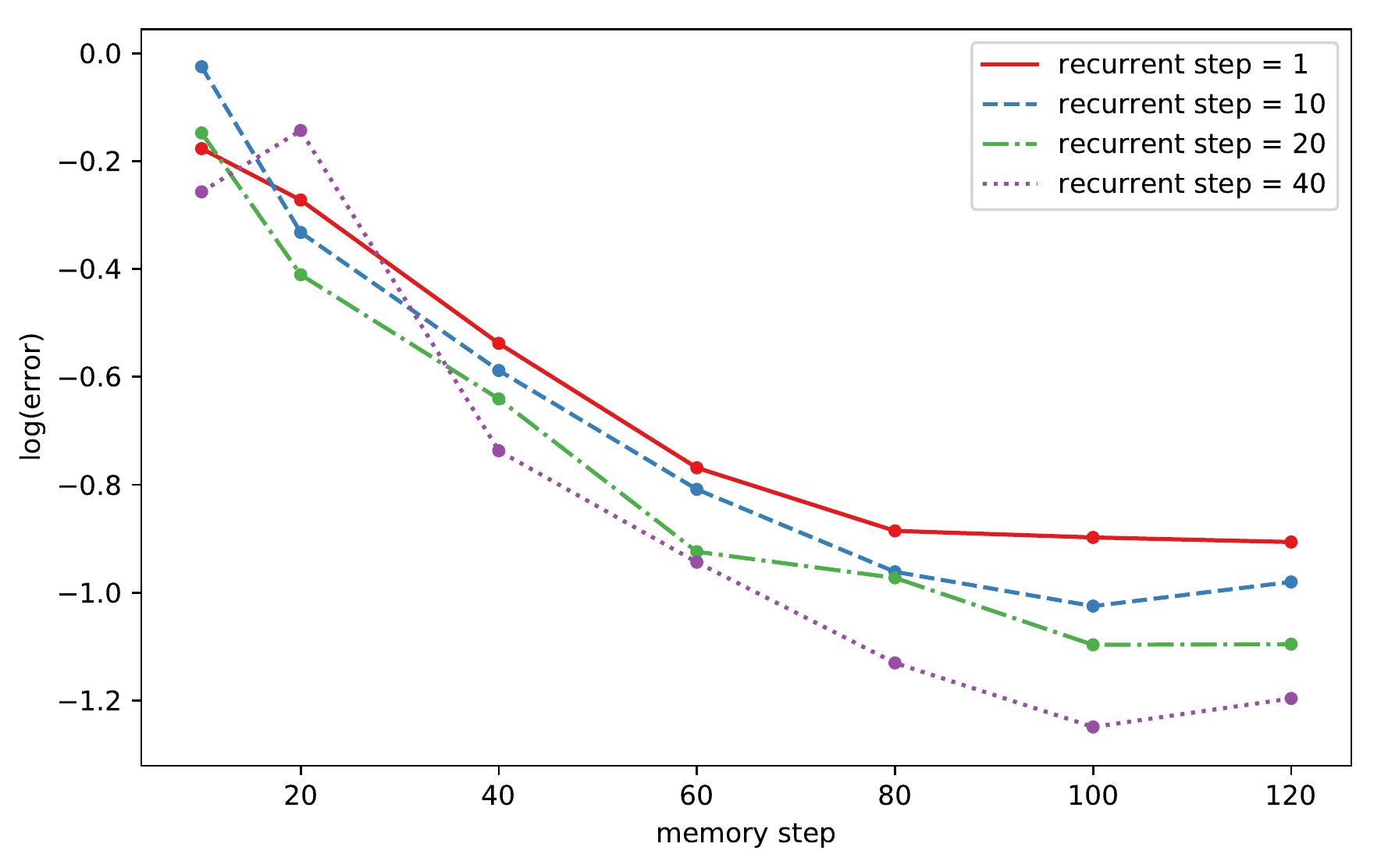}
		\caption{Errors {\em vs.} $n_M$.}
		\label{fig:eg1_memory_error}
	\end{subfigure}%
	\begin{subfigure}[b]{0.5\textwidth}
		\includegraphics[width=\textwidth]{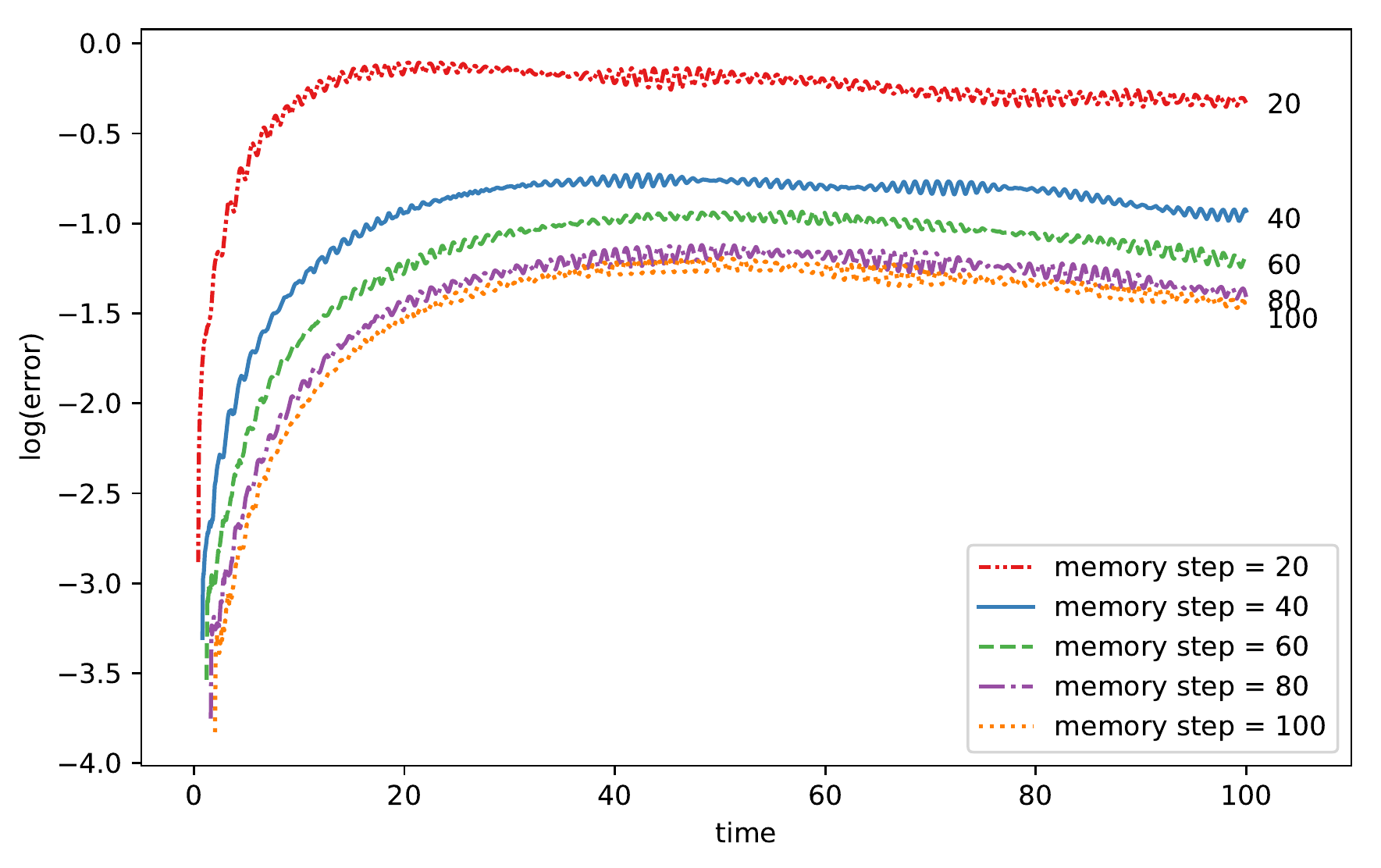}
		\caption{Errors over time for $n_R = 40$.}
		\label{fig:eg1_time_error}
	\end{subfigure}%
	\caption{Example 1. Model prediction errors at different
          memory steps and recurrent steps.
          }
	\label{fig:eg1_error}
\end{figure}

The memory step is tested for $n_M = 10, 20, 40, 60, 80, 100, 120$,
and the recurrent step is tested for $n_R = 1, 10, 20, 40$. The model
prediction errors at different memory steps and recurrent steps are
shown in Fig.~\ref{fig:eg1_error}.
The prediction errors are computed using $\ell_2$-norm of the DNN
model predictions against the reference solutions at time level $t=100$, averaged over 100
simulations with random initial conditions and system parameters.
We observe that
the accuracy improvement over increasing $n_M$ starts to saturate with
$n_M\geq 100$. We also notice that a larger $n_R$ produces better
results consistently. 

\begin{figure}
	\centering
	\begin{subfigure}[b]{0.5\textwidth}
		\includegraphics[width=\textwidth]{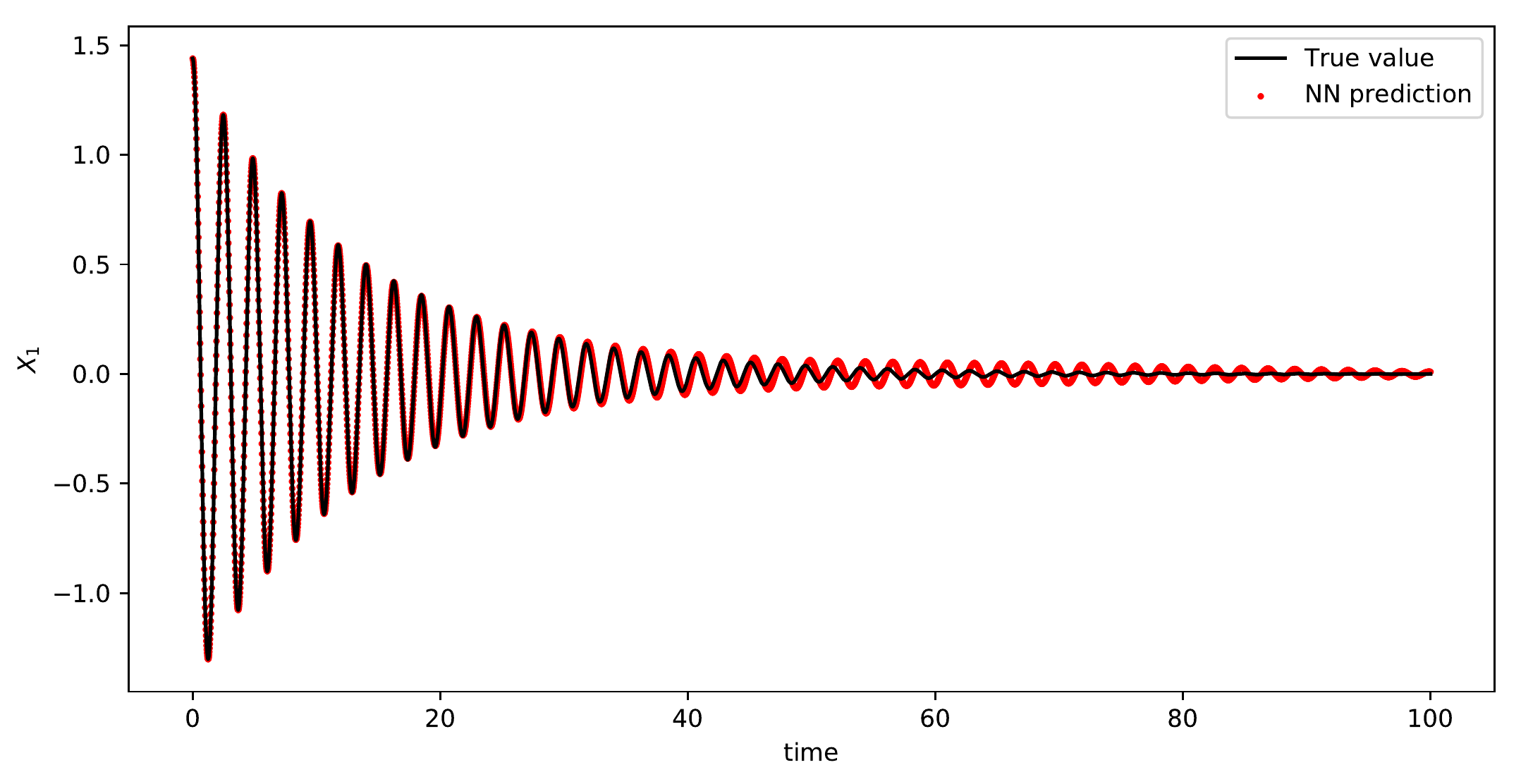}
		\caption{$x_1$}
	\end{subfigure}%
	\begin{subfigure}[b]{0.5\textwidth}
		\includegraphics[width=\textwidth]{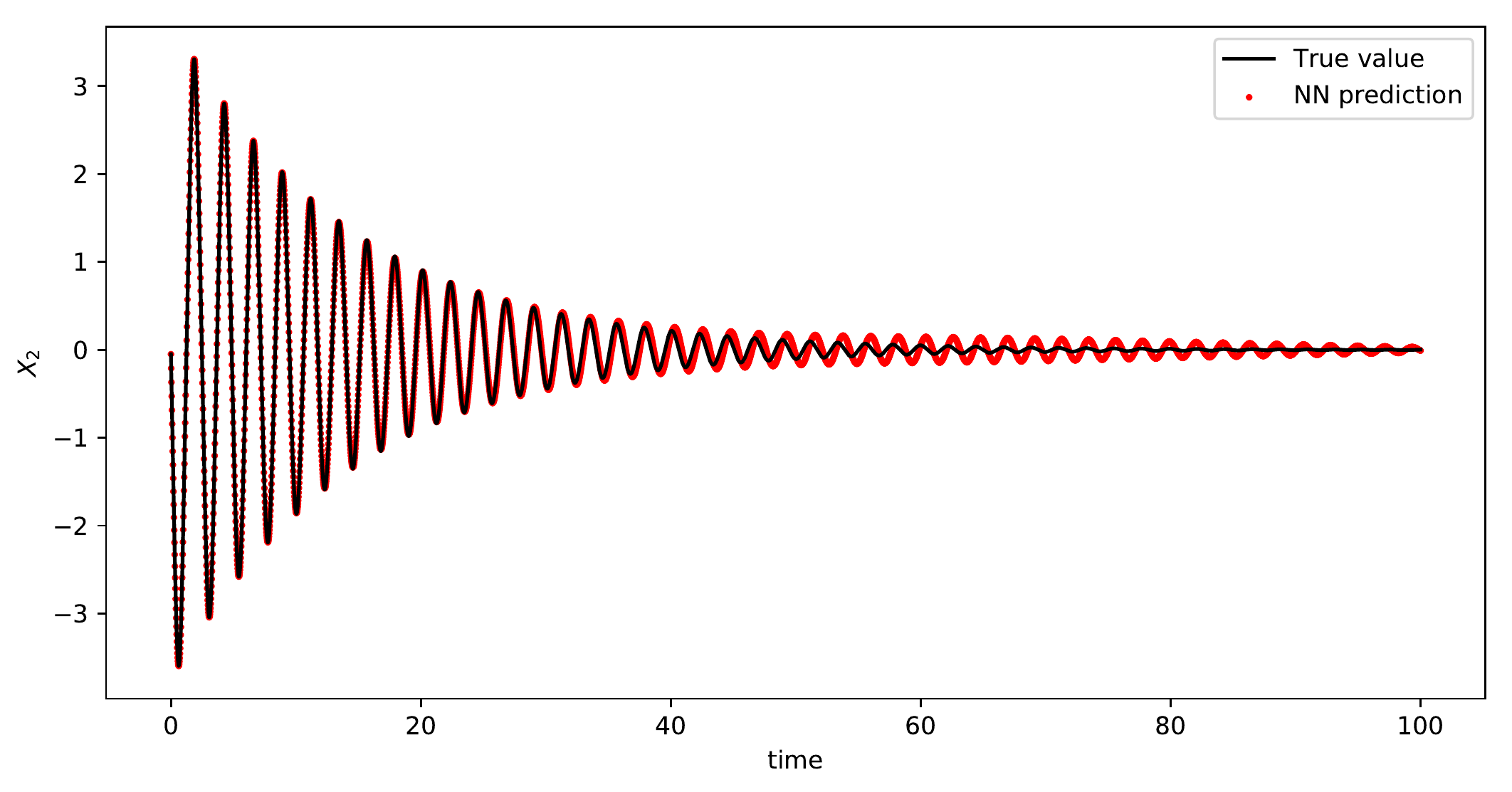}
		\caption{$x_2$}
	\end{subfigure}%
\hfill
\begin{subfigure}[b]{0.5\textwidth}
	\includegraphics[width=\textwidth]{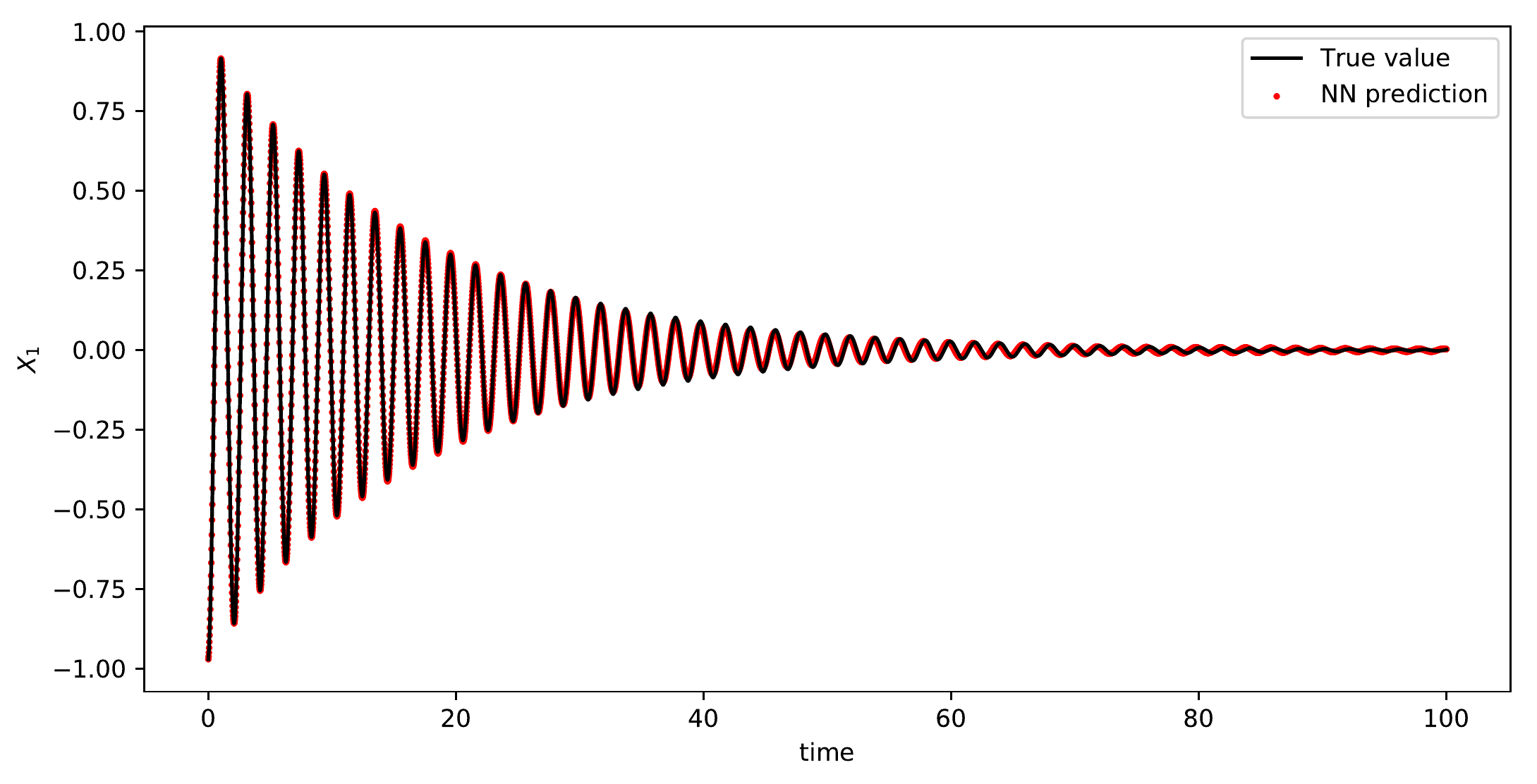}
	\caption{$x_1$}
\end{subfigure}%
\begin{subfigure}[b]{0.5\textwidth}
	\includegraphics[width=\textwidth]{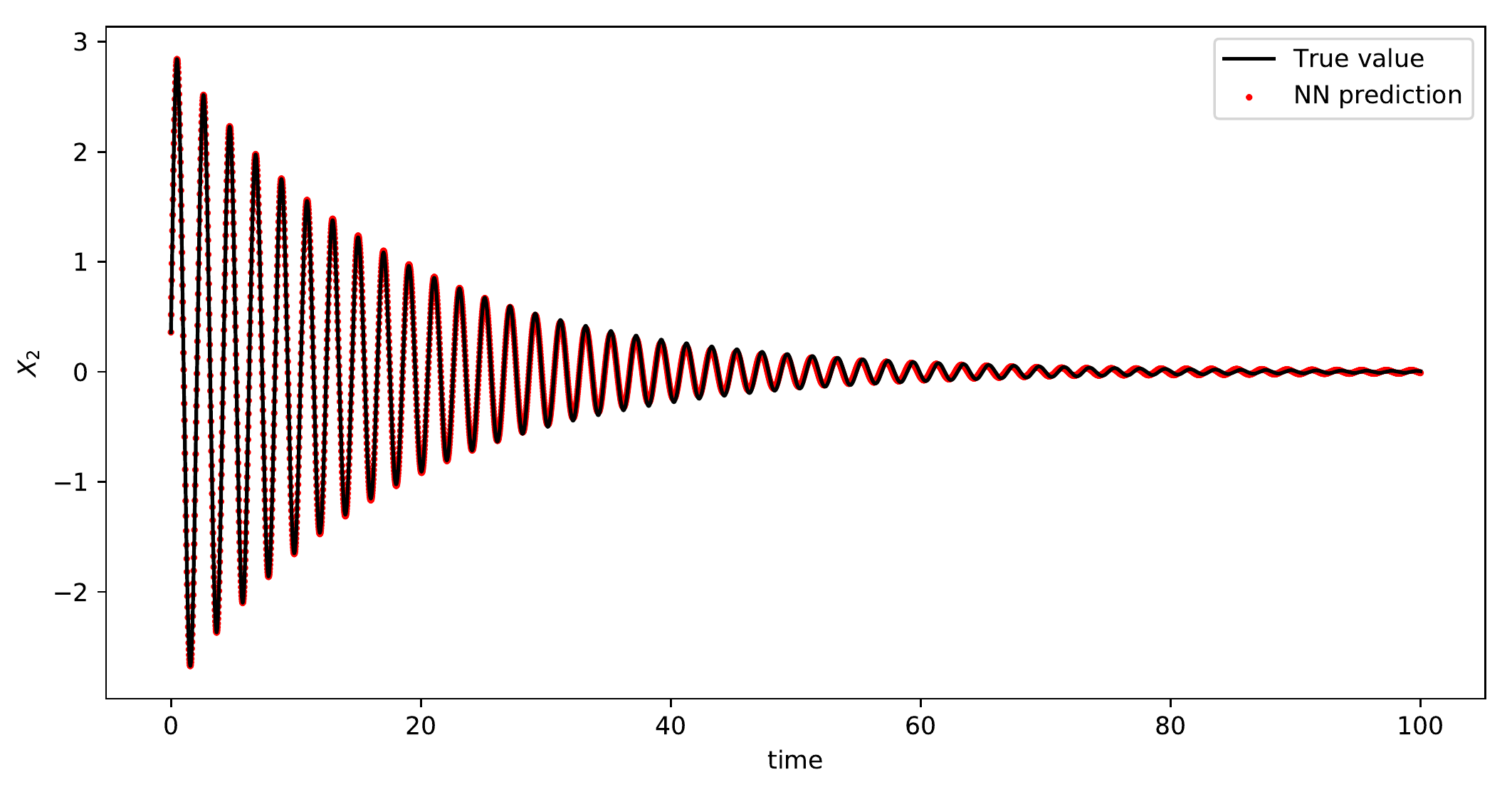}
	\caption{$x_2$}
\end{subfigure}%
	\caption{Example 1. Model predictions up to $t=100$ with
          $n_M=100$ and $n_R=40$ using two sets of arbitrary initial
          conditions and system parameters.}
	\label{fig:eg1_prediction}
\end{figure}

\begin{figure}
	\centering
	\begin{subfigure}[b]{0.34\textwidth}
		\includegraphics[width=\textwidth]{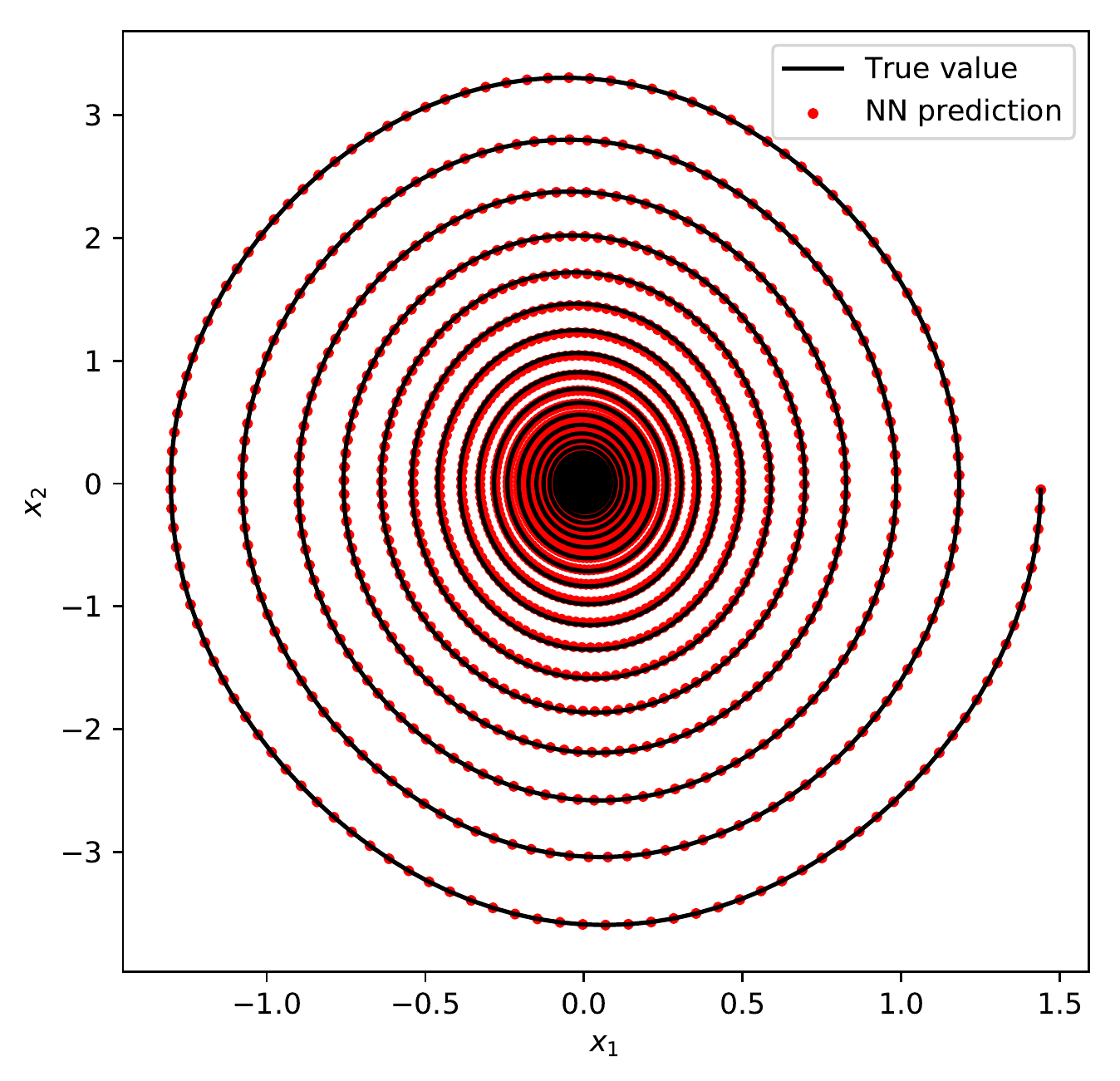}
		\caption{phase plot}
	\end{subfigure}%
	\begin{subfigure}[b]{0.64\textwidth}
		\includegraphics[width=\textwidth]{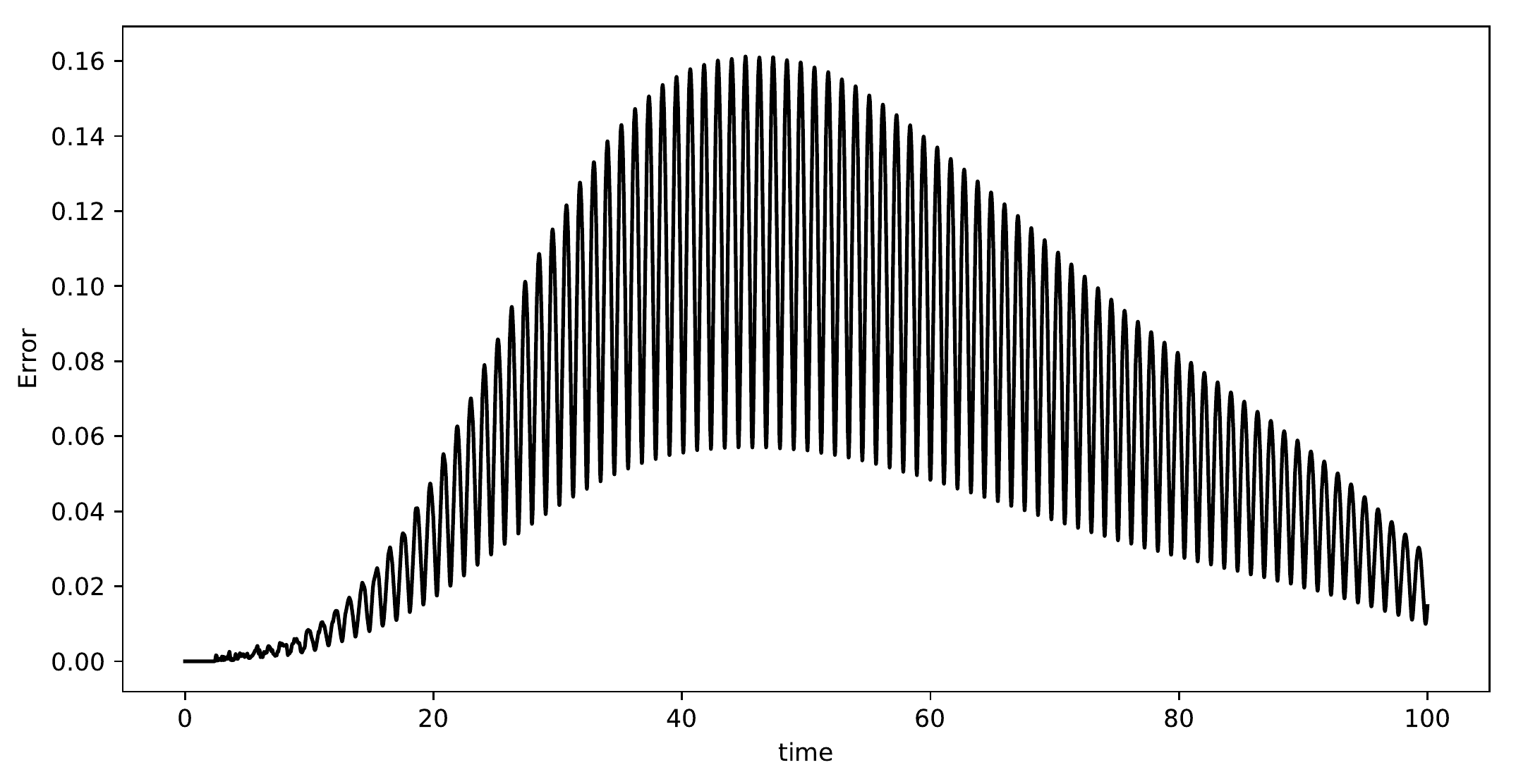}
		\caption{error}
	\end{subfigure}%
	\hfill
	\begin{subfigure}[b]{0.34\textwidth}
		\includegraphics[width=\textwidth]{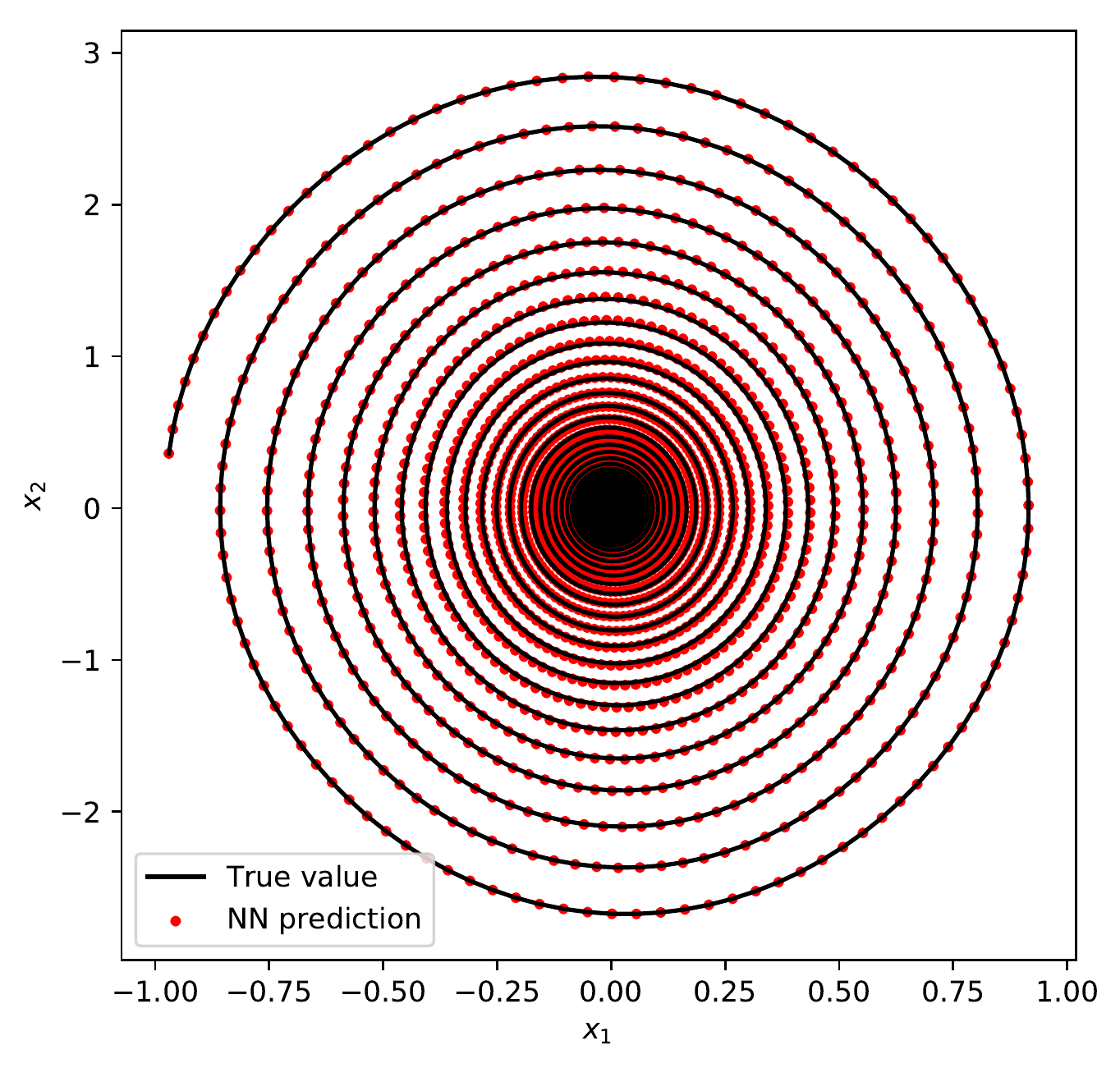}
		\caption{phase plot}
	\end{subfigure}%
	\begin{subfigure}[b]{0.64\textwidth}
		\includegraphics[width=\textwidth]{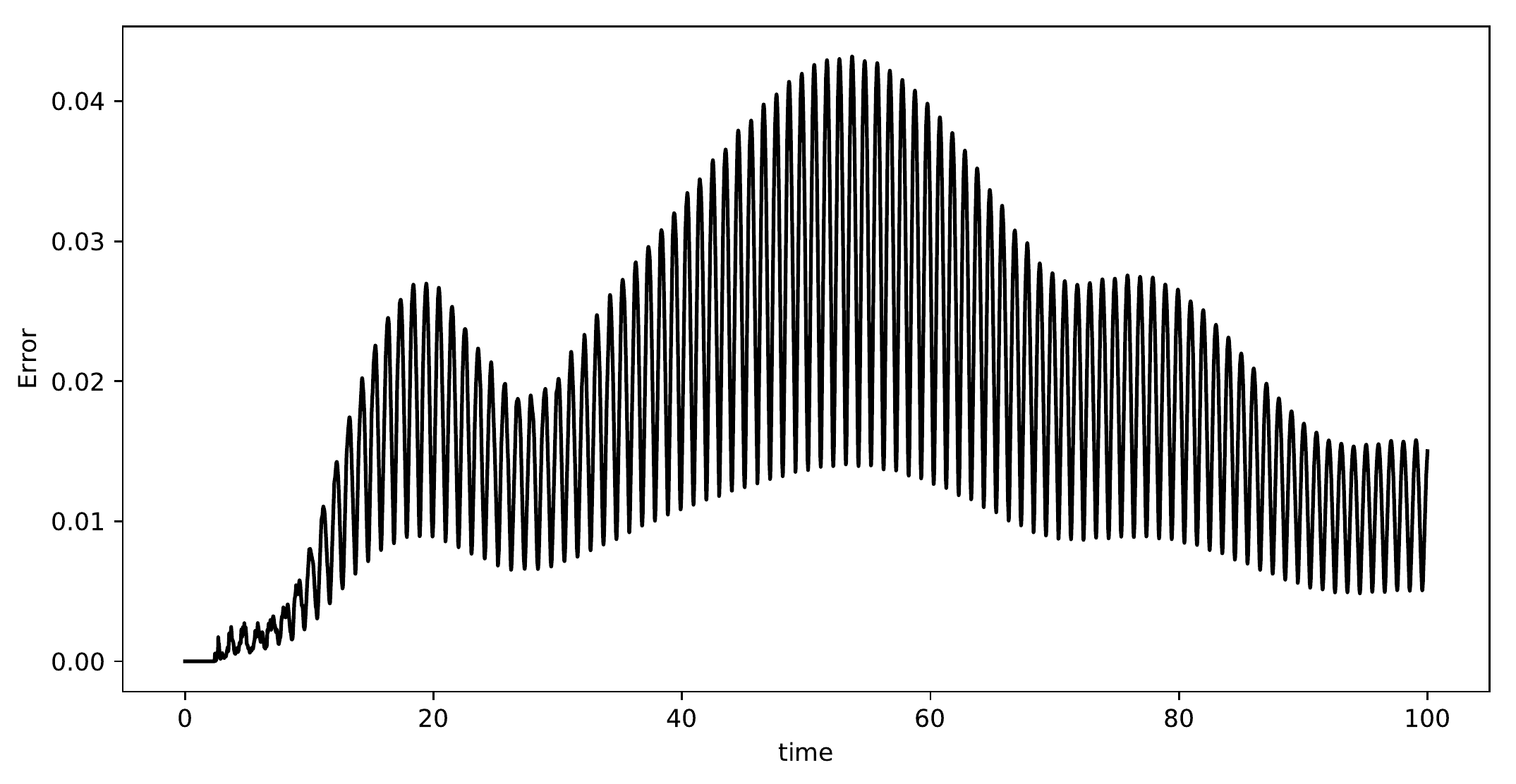}
		\caption{error}
	\end{subfigure}%
	\caption{Example 1. Model predictions and errors up to $t=100$
          with $n_M=100$ and $n_R=40$ using two sets of arbitrary
          initial conditions and parameters as in Fig.~\ref{fig:eg1_prediction}.
          }
	\label{fig:eg1_phase}
\end{figure}

The DNN model predictive results with $n_M=100$ and $n_R=40$ are shown in
Fig.~\ref{fig:eg1_prediction}, with two sets of arbitrarily chosen
initial conditions and (hidden) system parameters. This corresponds to memory length
$n_M\times \Delta = 0.4$, which is in fact rather short. We observe a
very good agreement between the DNN model predictions and the
reference solutions for the long-term integrations up to $t=100$.  The
corresponding numerical errors are plotted in
Fig.~\ref{fig:eg1_phase}, along with the comparison of the phase
portraits.

\subsection{Example 2: Larger Linear System} \label{sec:example2}

We now consider a larger linear system involving 20 state variables
$$
\dot \x = \A \x, \qquad \x\in \Rs^{20}, \quad \A\in \Rs^{20\times 20},
$$
where among the 400 entries of the coefficient matrix $\A$, we treat 100 of them as
hidden parameters. More specifically, let us rewrite the system in
term of
$\x=(\p; \q)$, where $\p\in\Rs^{10}$ and $\q\in\Rs^{10}$ satisfy
\begin{equation}
\begin{cases}
\dot \p = \Sigma_{11} \p + (\I + \Sigma_{12}) \q,\\
\dot \q = -(\I + \Sigma_{21}) \p - \Sigma_{22} \q.
\end{cases}       
\end{equation}
Here, $\I$ is the identity matrix of size $10\times
10$, and $\Sigma_{ij} \in \Rs^{10\times 10}$,
$i=1,2,j=1,2$ are four coefficient matrices. We set three of the
coefficient matrices to be known, with $\Sigma_{11} = \Sigma_{12} = \mathbf{0}$, and
$\Sigma_{22}$ with the 
entries listed in the Appendix. The 100 entries of the matrix $\Sigma_{21}$ 
are treated as hidden parameters within the domain $[-0.05,0.05]^{100}$.
The domain of interest for the state variables is set as
$[-2,2]^{20}$.

\begin{figure}
	\centering
	\begin{subfigure}[b]{0.5\textwidth}
		\includegraphics[width=\textwidth]{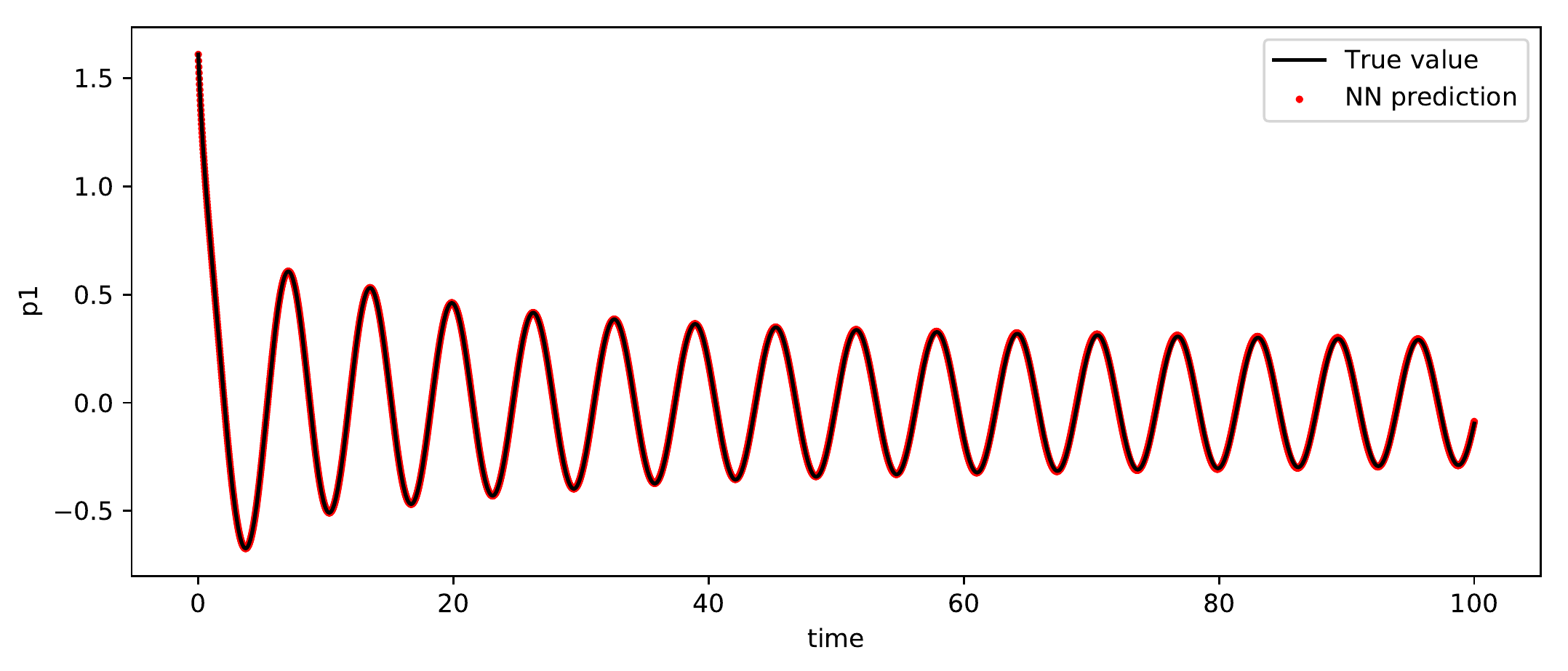}
	\end{subfigure}%
	\begin{subfigure}[b]{0.5\textwidth}
		\includegraphics[width=\textwidth]{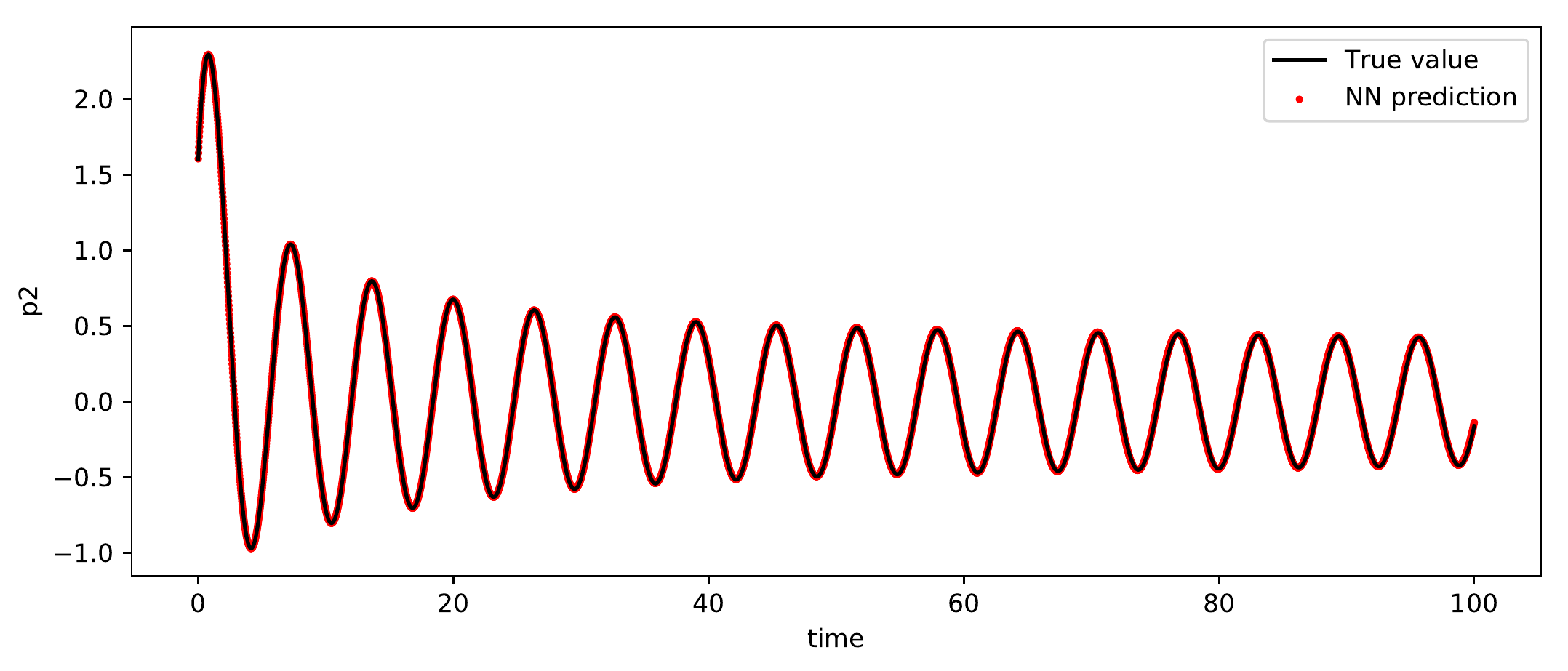}
	\end{subfigure}%
	\hfill
	\begin{subfigure}[b]{0.5\textwidth}
		\includegraphics[width=\textwidth]{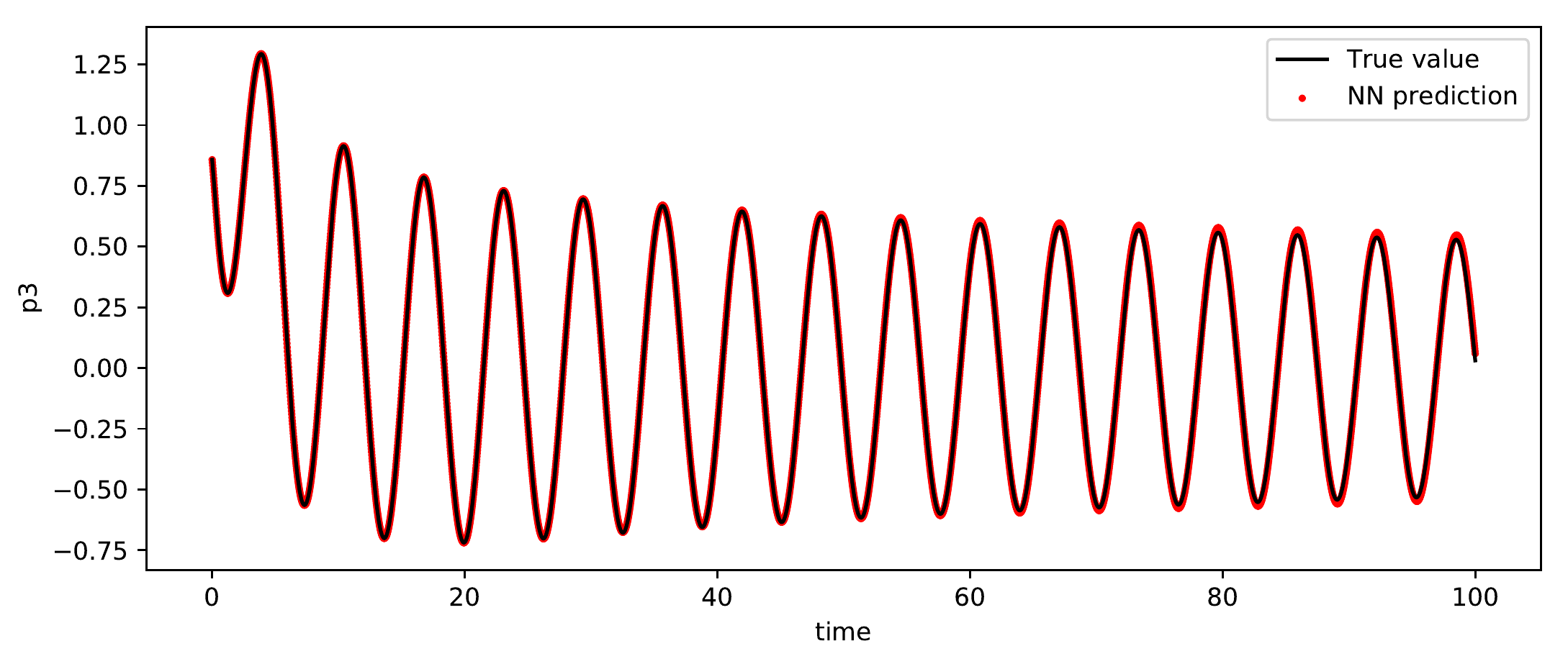}
	\end{subfigure}%
	\begin{subfigure}[b]{0.5\textwidth}
		\includegraphics[width=\textwidth]{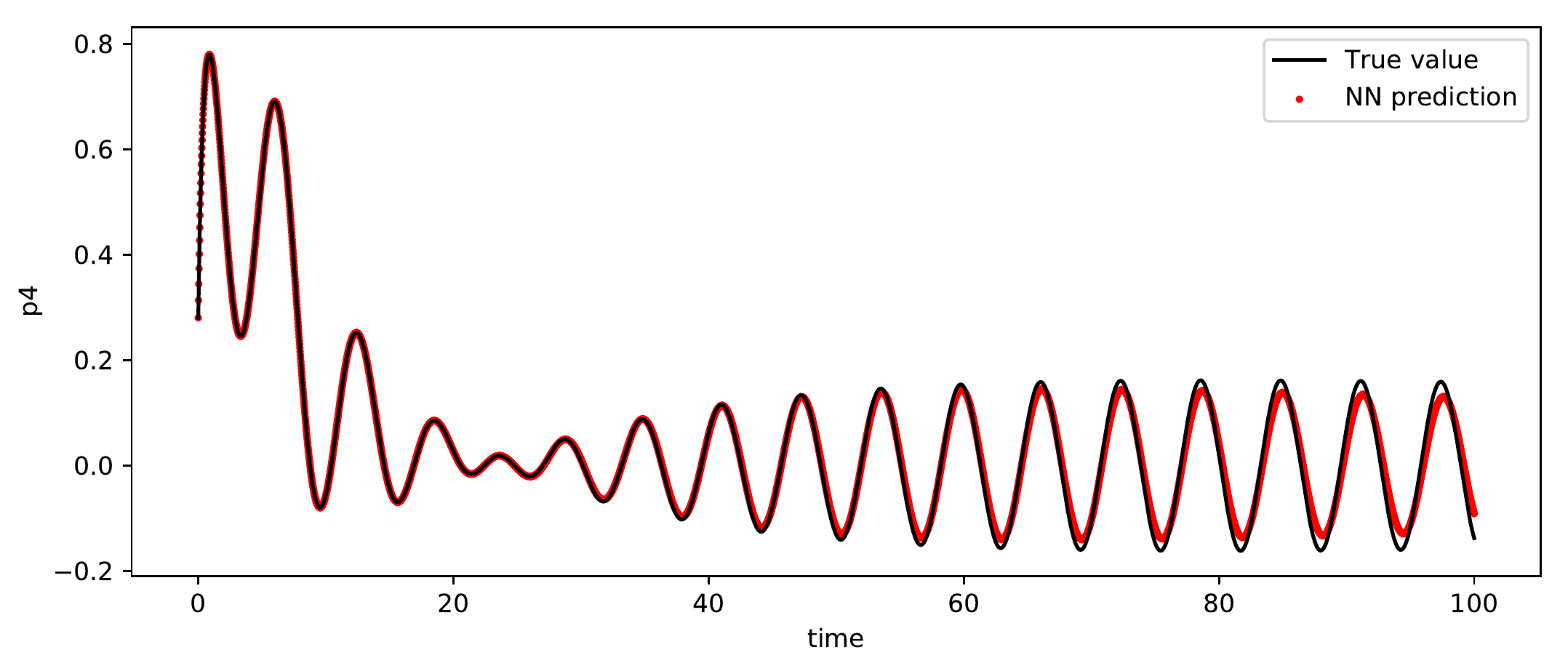}
	\end{subfigure}%
	\hfill
	\begin{subfigure}[b]{0.5\textwidth}
		\includegraphics[width=\textwidth]{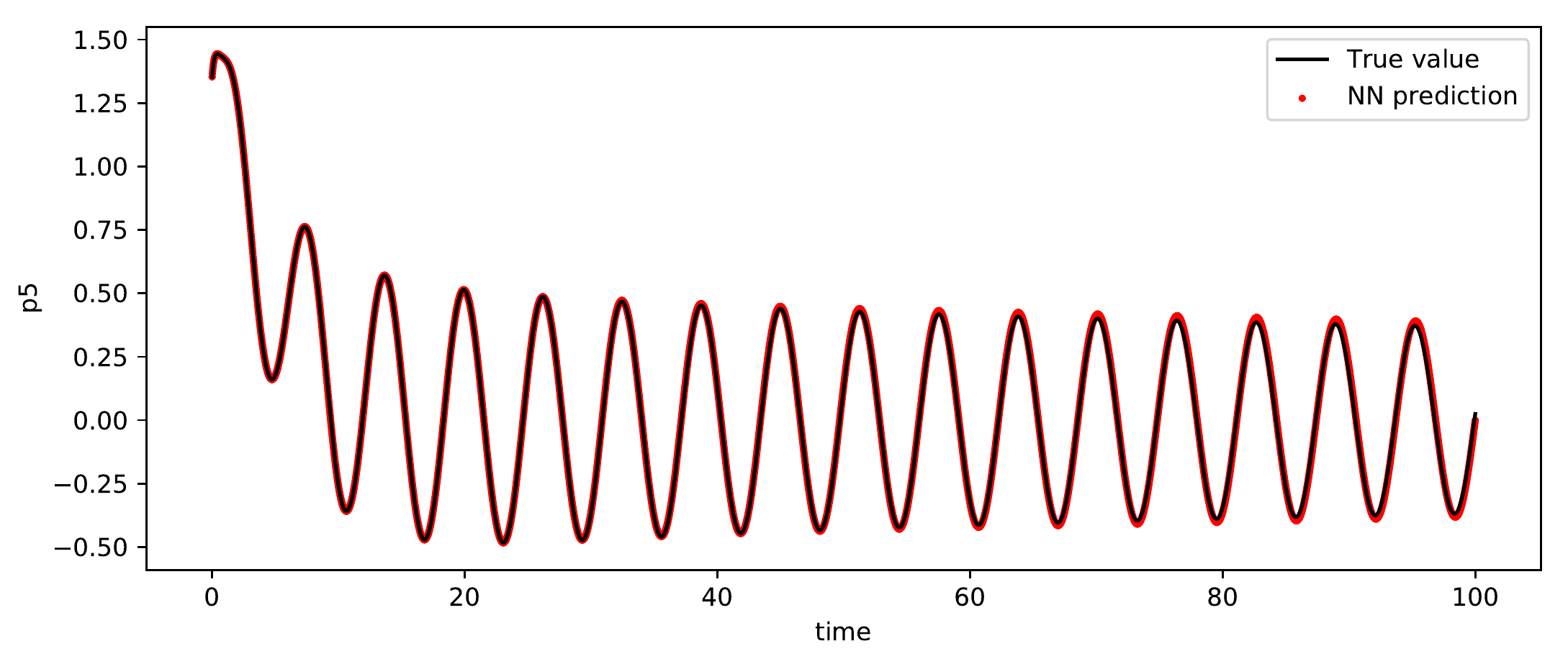}
	\end{subfigure}%
	\begin{subfigure}[b]{0.5\textwidth}
		\includegraphics[width=\textwidth]{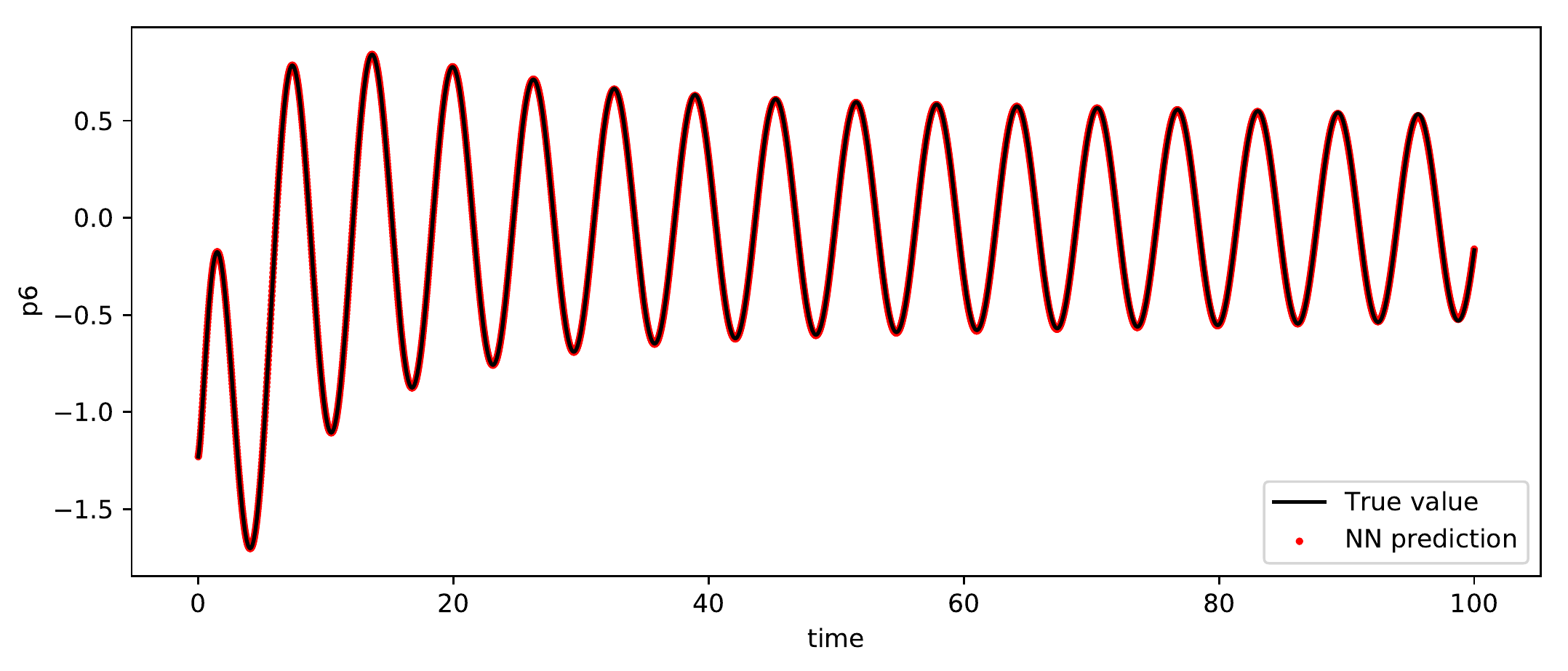}
	\end{subfigure}%
	\hfill
	\begin{subfigure}[b]{0.5\textwidth}
		\includegraphics[width=\textwidth]{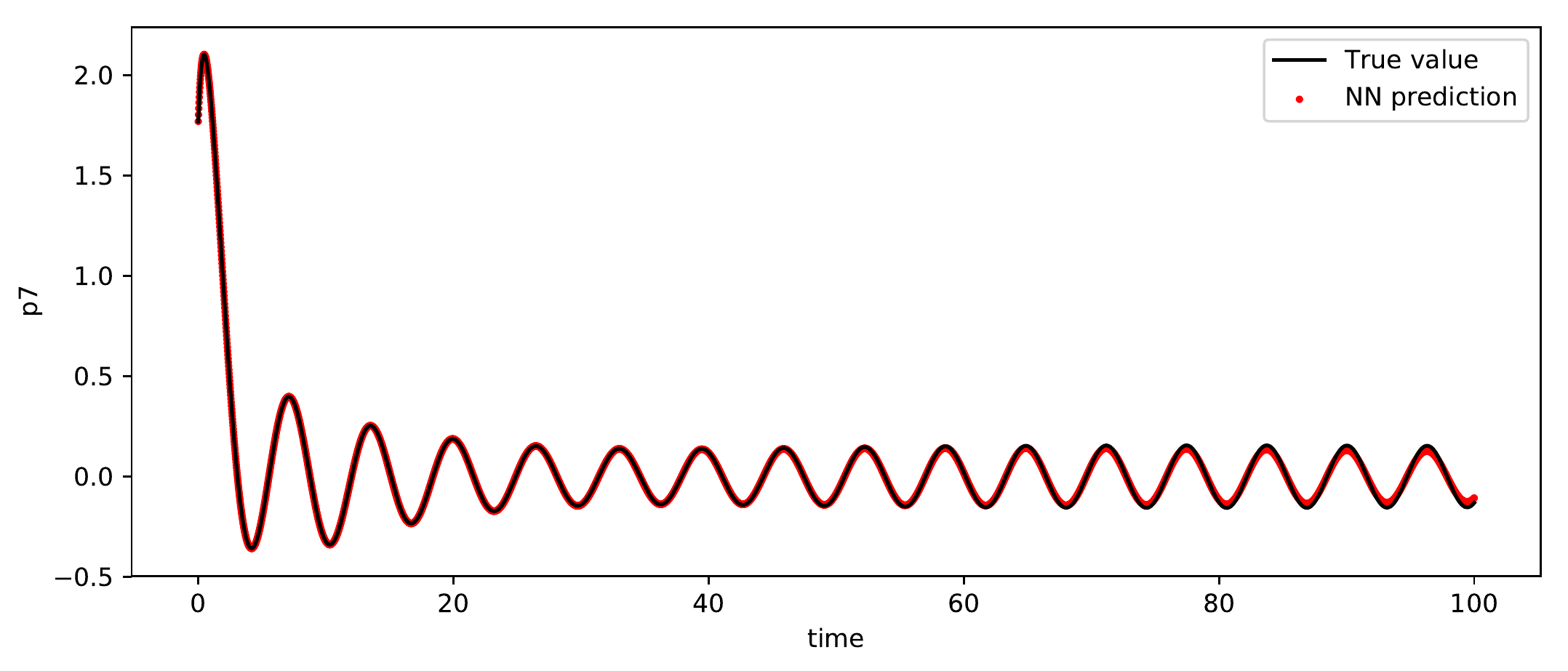}
	\end{subfigure}%
	\begin{subfigure}[b]{0.5\textwidth}
		\includegraphics[width=\textwidth]{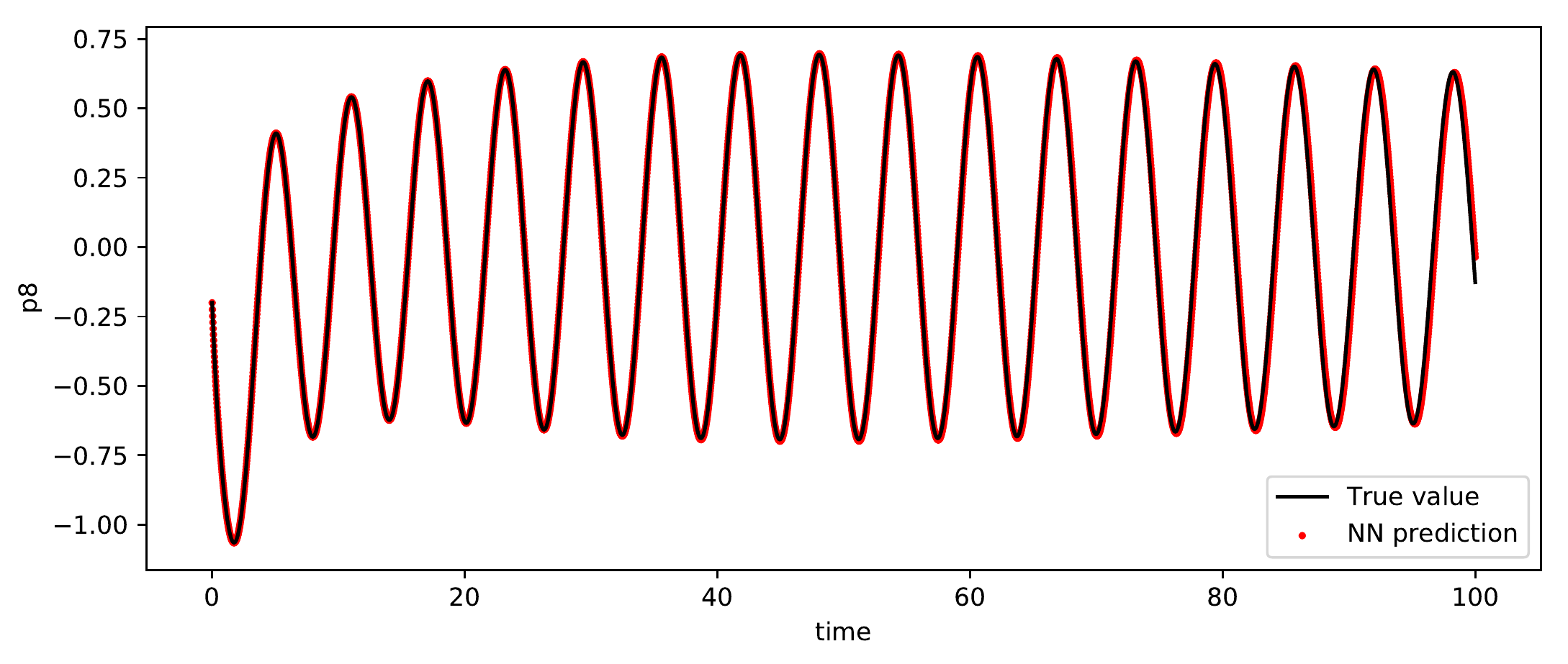}
	\end{subfigure}%
	\hfill
	\begin{subfigure}[b]{0.5\textwidth}
		\includegraphics[width=\textwidth]{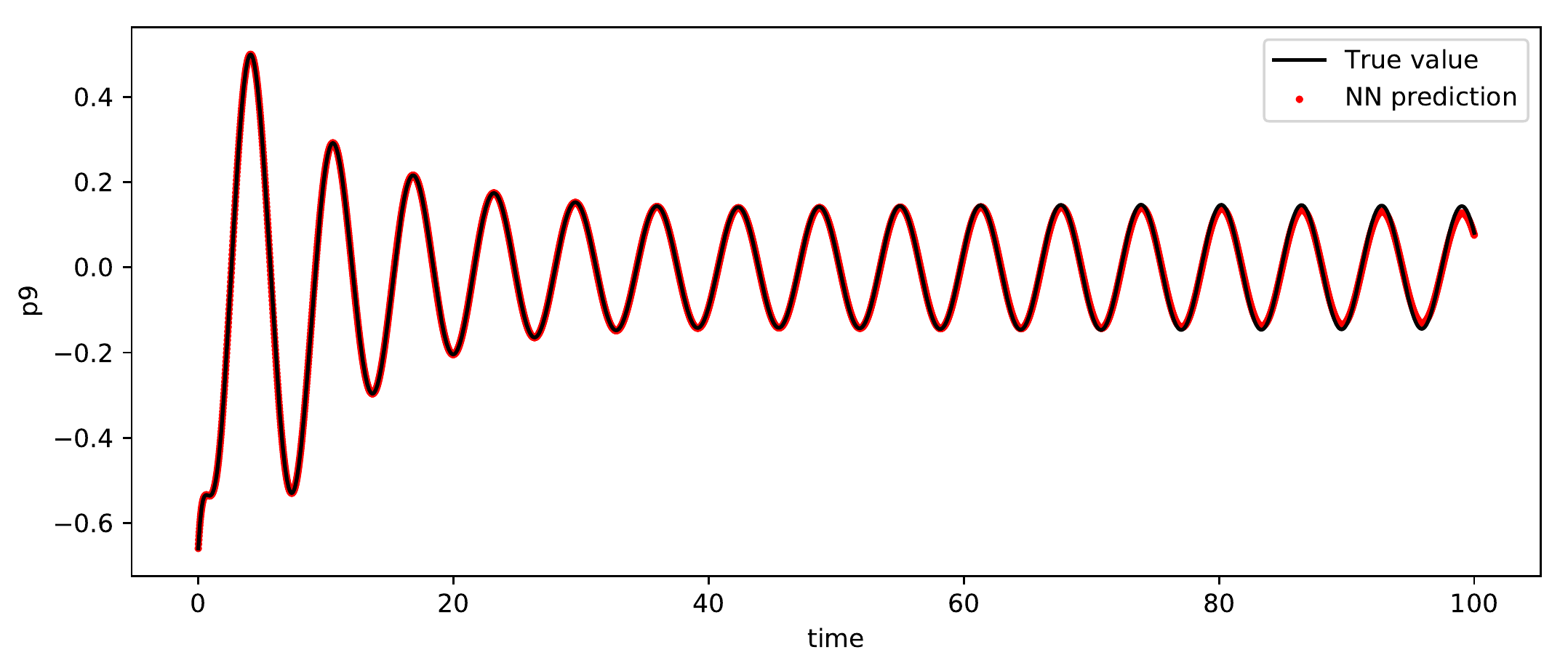}
	\end{subfigure}%
	\begin{subfigure}[b]{0.5\textwidth}
		\includegraphics[width=\textwidth]{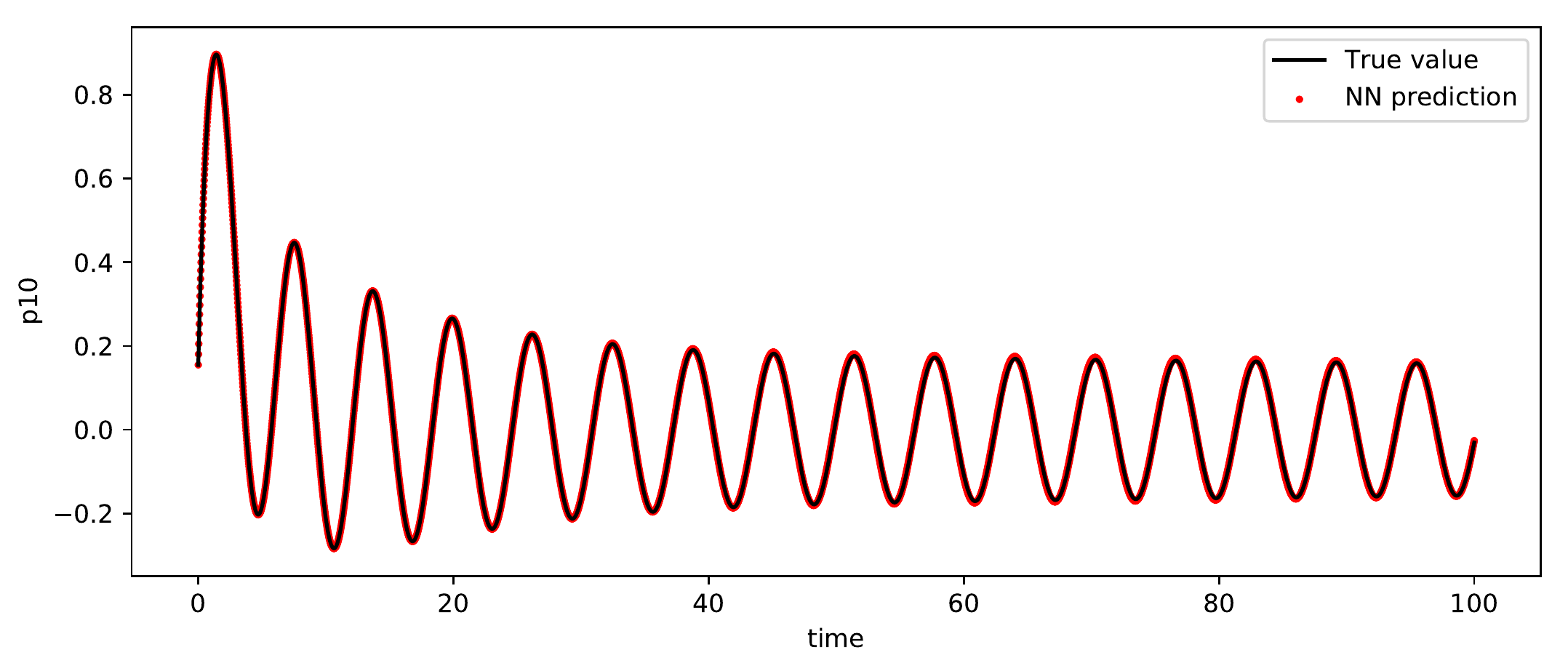}
	\end{subfigure}%
	\hfill
	\caption{Example 2. Model predictions of $\p$ up to $t=100$
          with $n_M=1,300$ and $n_R=1$.}
	\label{fig:eg2_prediction1}
\end{figure}

\begin{figure}
	\centering
	\begin{subfigure}[b]{0.5\textwidth}
		\includegraphics[width=\textwidth]{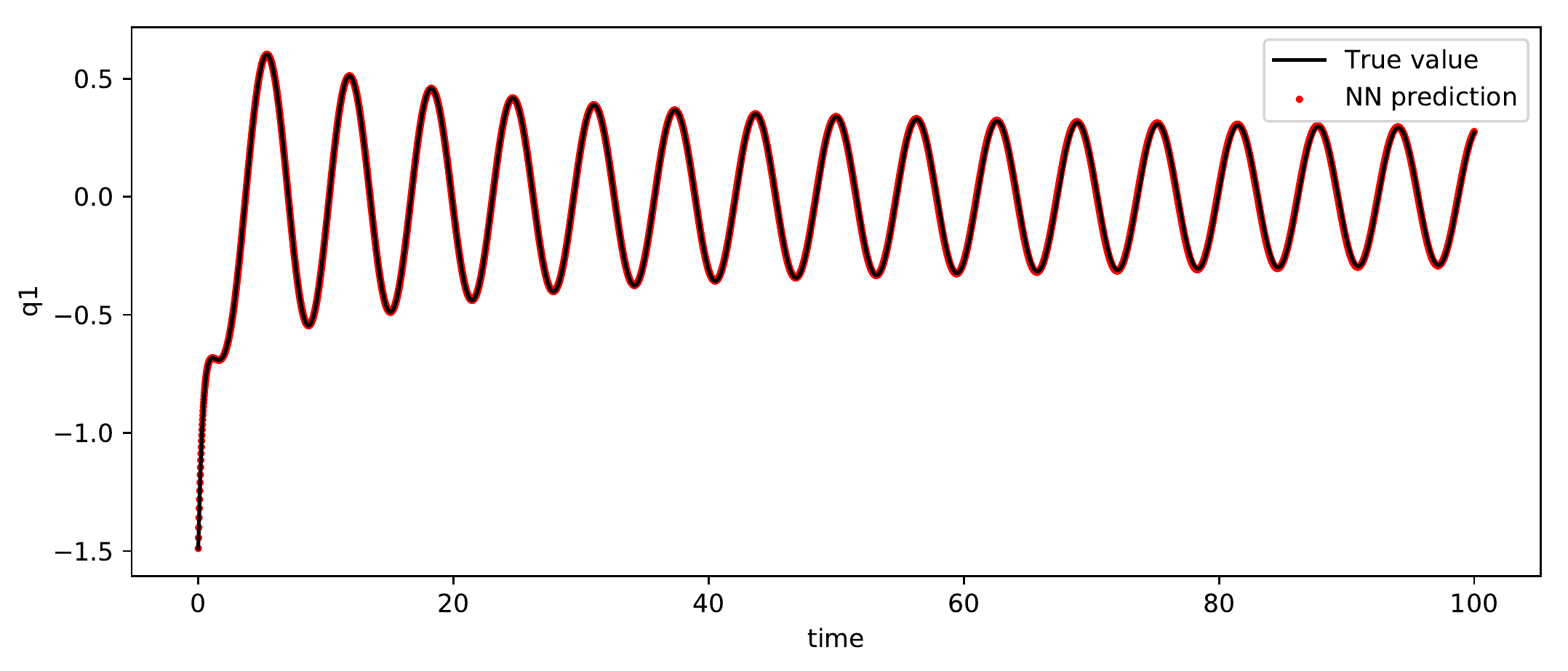}
	\end{subfigure}%
	\begin{subfigure}[b]{0.5\textwidth}
		\includegraphics[width=\textwidth]{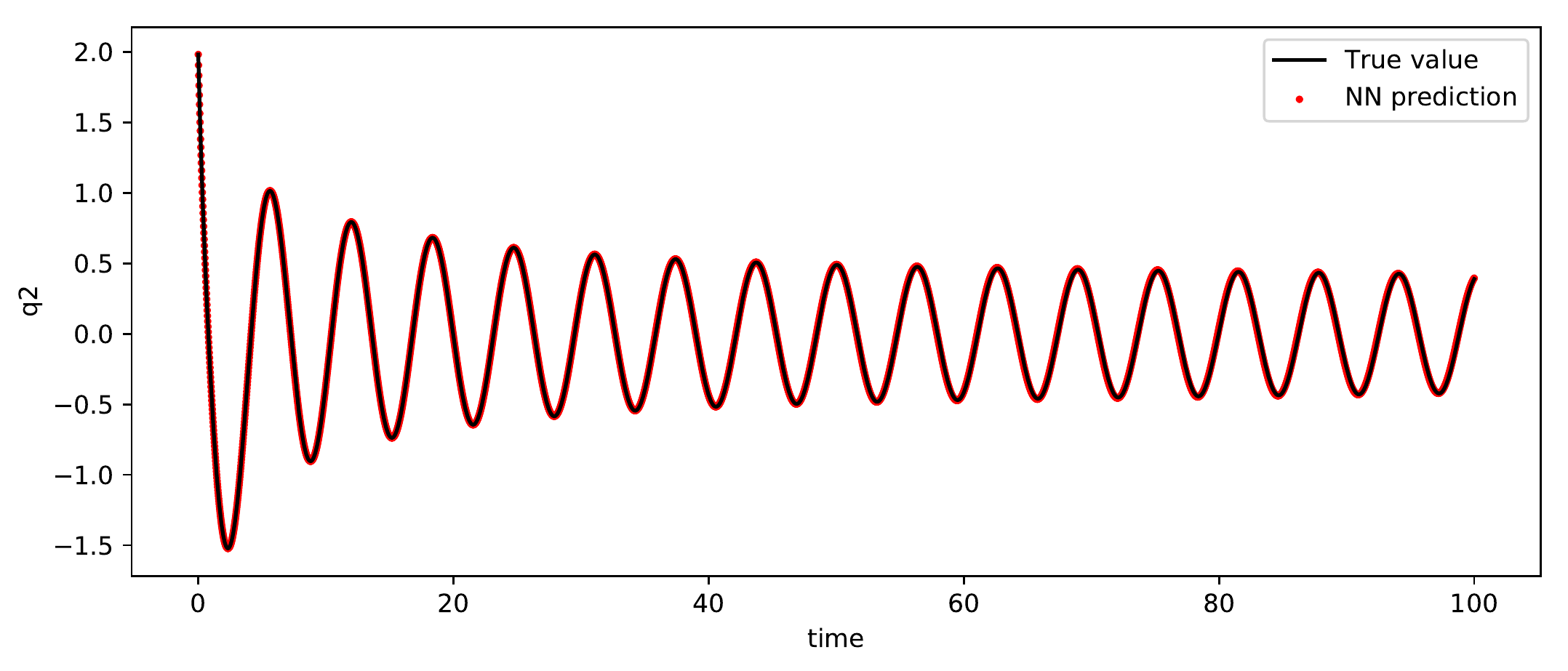}
	\end{subfigure}%
	\hfill
	\begin{subfigure}[b]{0.5\textwidth}
		\includegraphics[width=\textwidth]{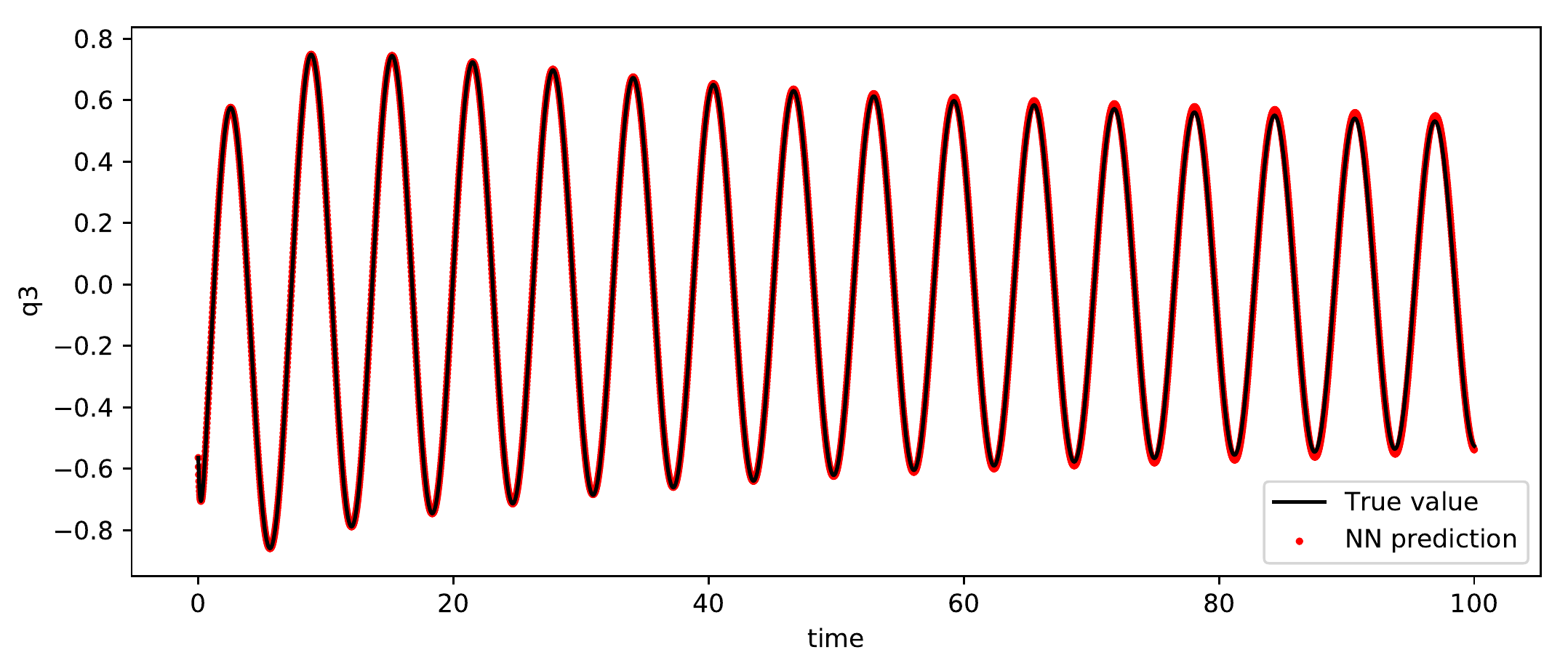}
	\end{subfigure}%
	\begin{subfigure}[b]{0.5\textwidth}
		\includegraphics[width=\textwidth]{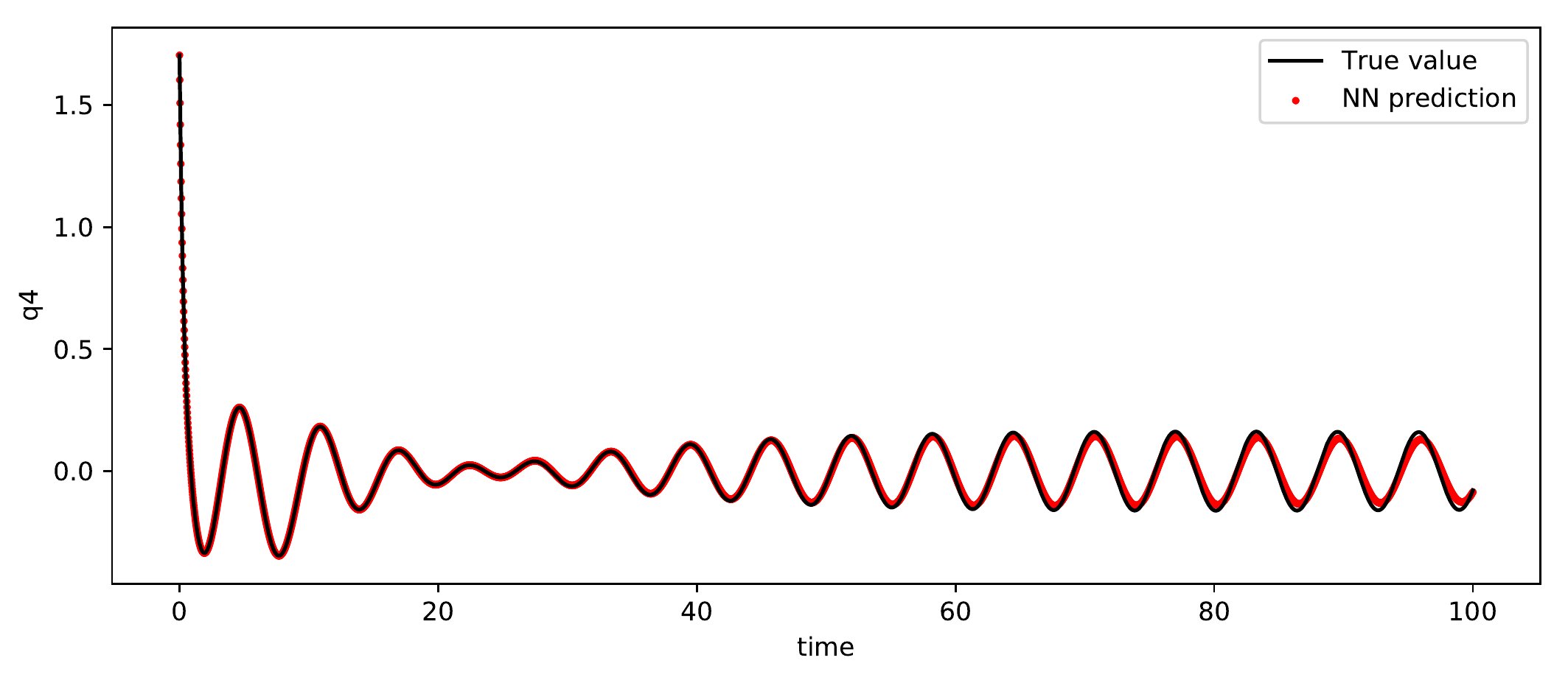}
	\end{subfigure}%
	\hfill
	\begin{subfigure}[b]{0.5\textwidth}
		\includegraphics[width=\textwidth]{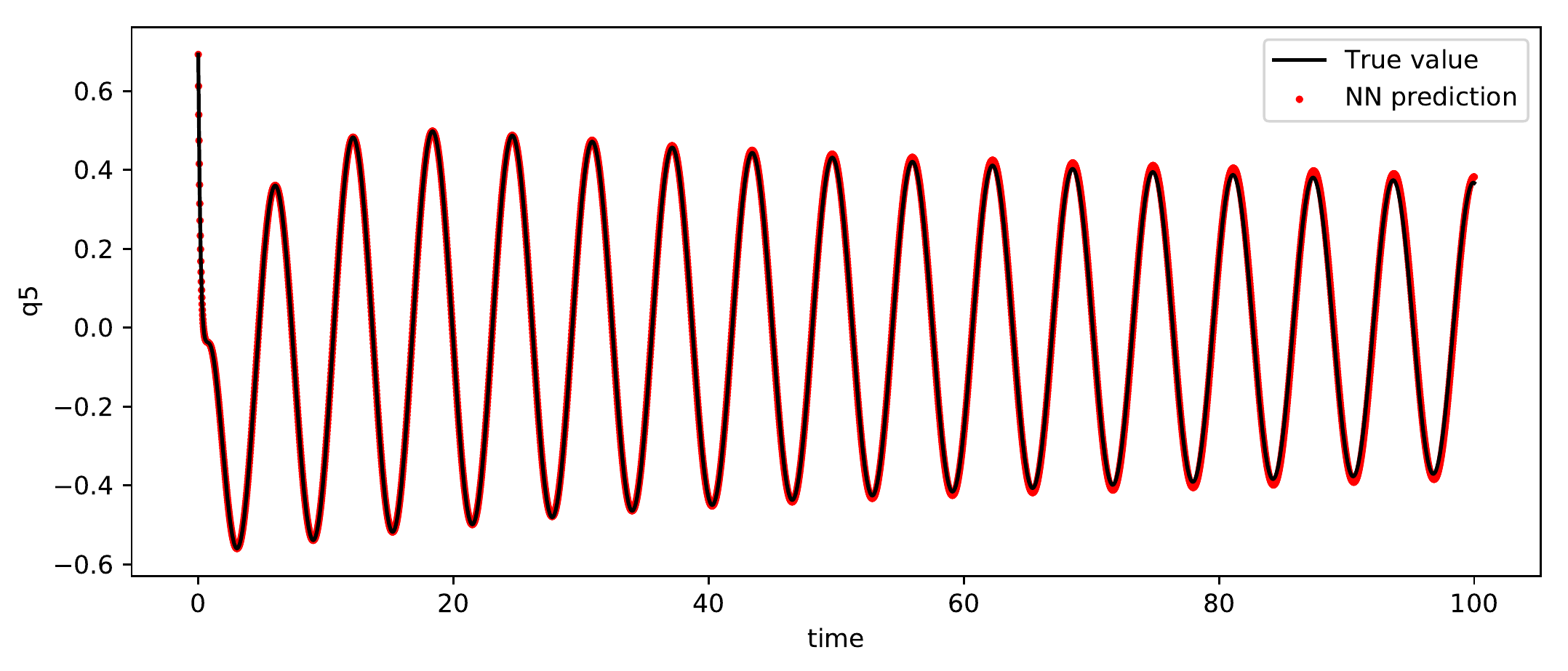}
	\end{subfigure}%
	\begin{subfigure}[b]{0.5\textwidth}
		\includegraphics[width=\textwidth]{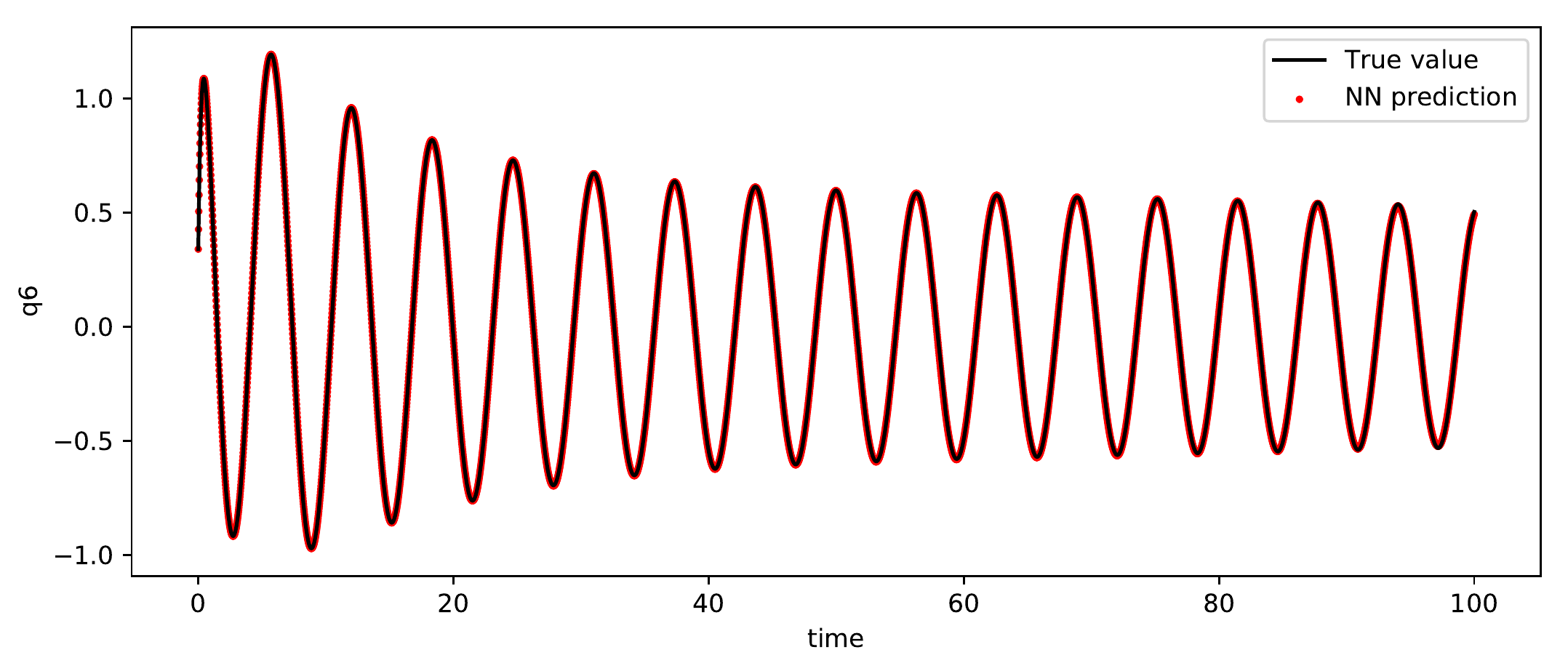}
	\end{subfigure}%
	\hfill
	\begin{subfigure}[b]{0.5\textwidth}
		\includegraphics[width=\textwidth]{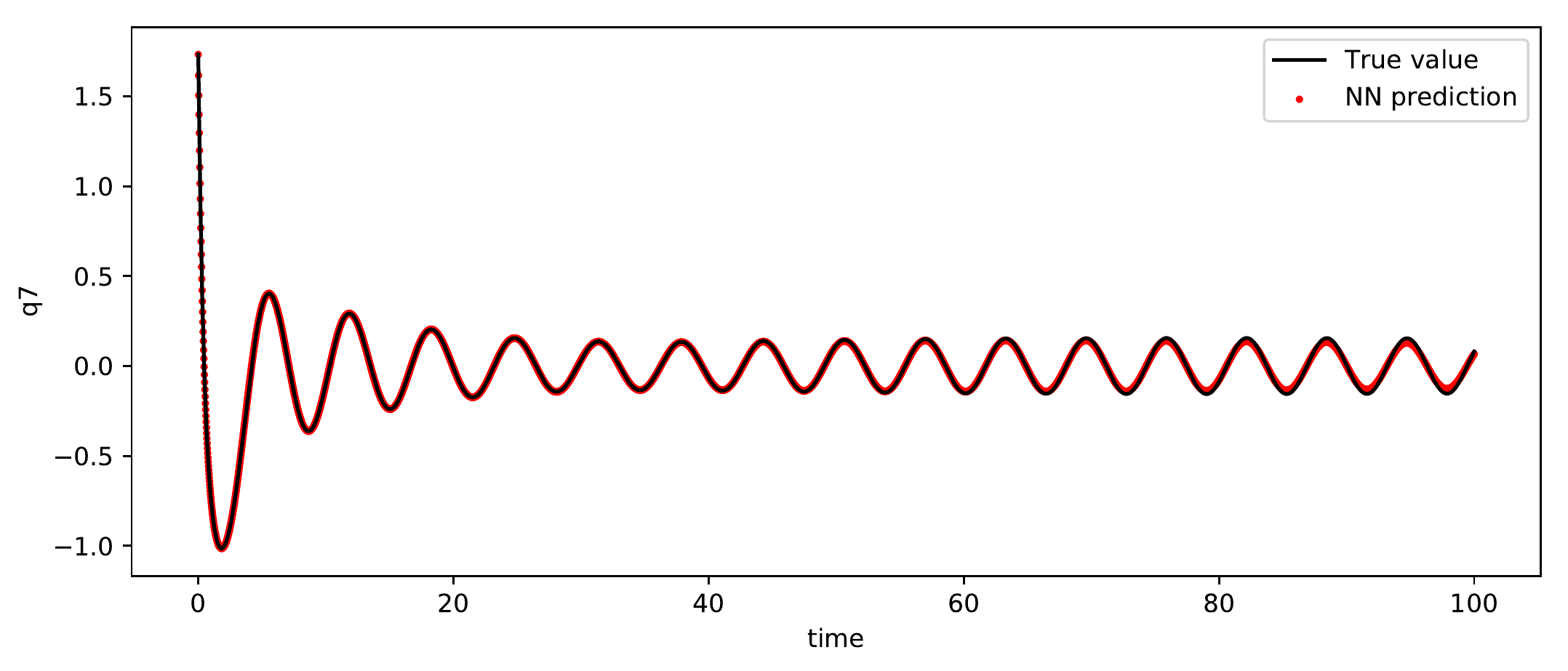}
	\end{subfigure}%
	\begin{subfigure}[b]{0.5\textwidth}
		\includegraphics[width=\textwidth]{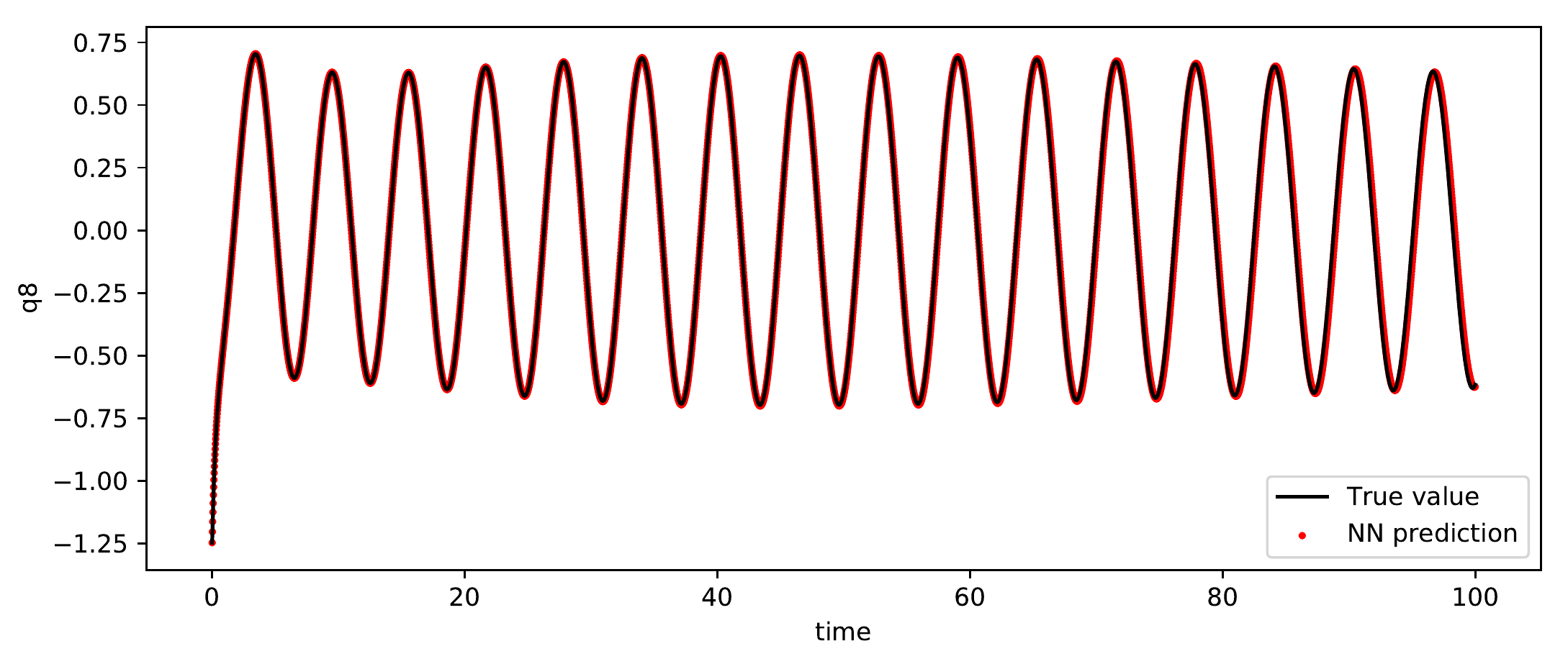}
	\end{subfigure}%
	\hfill
	\begin{subfigure}[b]{0.5\textwidth}
		\includegraphics[width=\textwidth]{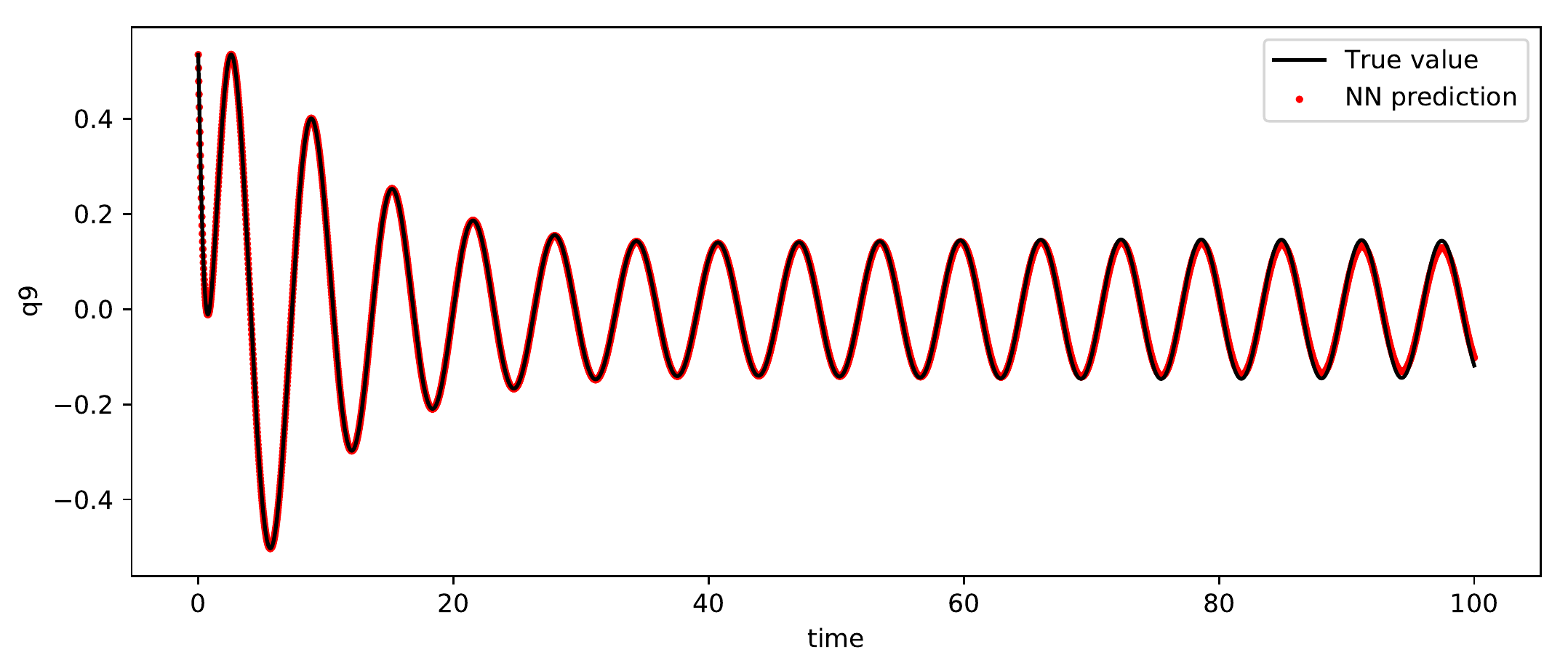}
	\end{subfigure}%
	\begin{subfigure}[b]{0.5\textwidth}
		\includegraphics[width=\textwidth]{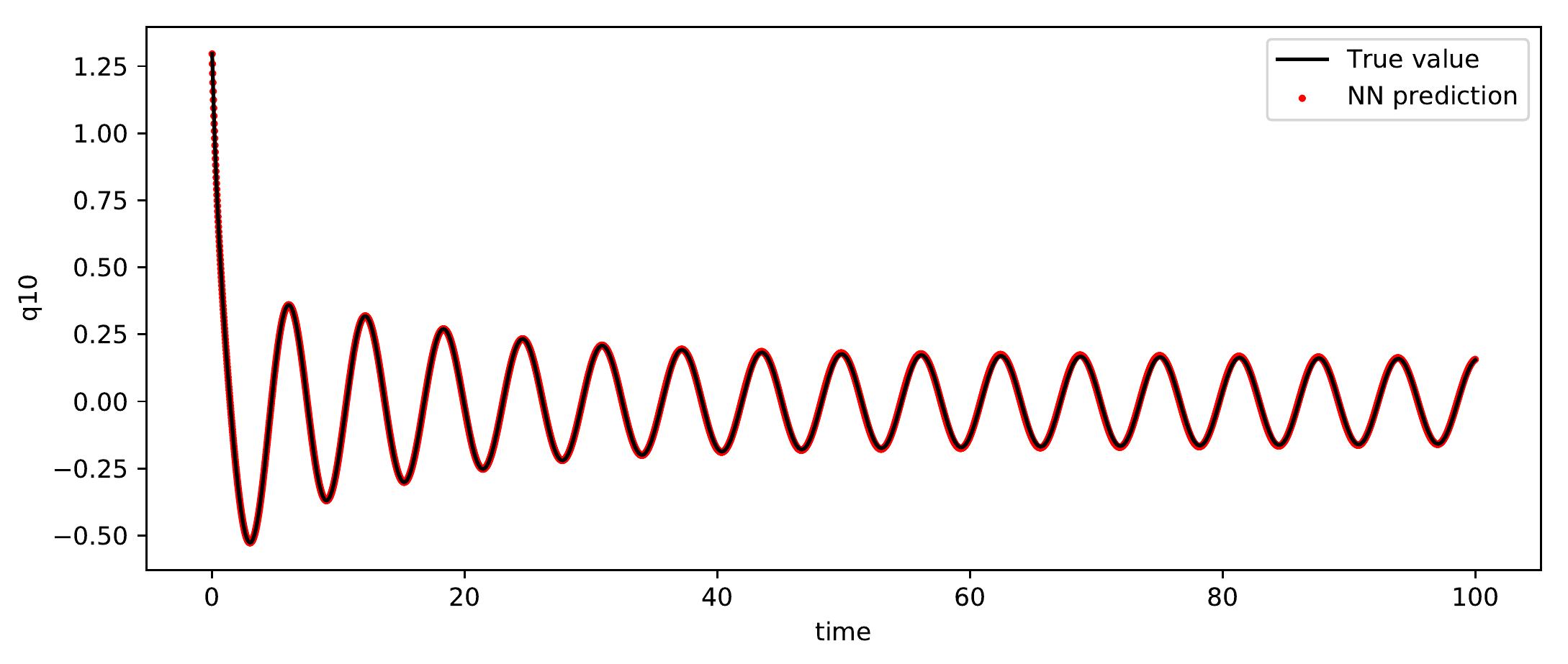}
	\end{subfigure}%
	\hfill
	\caption{Example 2. Model predictions of $\q$ up to $t=100$ with $n_M=1,300$ and $n_R=1$.}
	\label{fig:eg2_prediction2}
\end{figure}

With a larger number of missing hidden parameters (compared to Example
1), this problem requires longer memory length to construct an
accurate DNN model.
Memory steps of $n_M=100$, $300$, $500$, $700$, $900$, $1100$,
$1,300$, and $1,500$ are tested. The results indicate the $n_M=1,300$
is sufficient to produce converged prediction results. The recurrent
step is tested for $n_R=1$ to $n_R=5$. For this problem, the number of
recurrent step does not induce noticeable difference in the
prediction. We therefore fix $n_R=1$. 
The DNN model predictions for long-term integration up to $t=100$ with
$n_M=1,300$ and $n_R=1$ are shown in Fig.~\ref{fig:eg2_prediction1}
for the state variables $\p$
and in Fig.~\ref{fig:eg2_prediction2} for the state variables $\q$, using a set of  
arbitrarily chosen initial conditions and hidden parameter values.
We observe very good agreement between the DNN model predictions and
the corresponding reference solutions.

\subsection{Example 3: CSTR}
We now consider a smaller nonlinear system with bifurcation behavior
controlled by the hidden parameter. It is a continuous stirred-tank
chemical reactor (CSTR) model with a single and irreversible
exothermic reaction. The (unknown) governing equations are
\begin{equation}\label{CSTR_govern}
	\begin{cases}
		\dot x_1 = -x_1+Da\cdot (1-x_1)\exp(\frac{x_2}{1+x_2/\gamma}),\\
		\dot x_2 = -x_2+B\cdot Da\cdot (1-x_1)\exp(\frac{x_2}{1+x_2/\gamma}) - \beta(x_2-x_{2c}),
	\end{cases}       
\end{equation}
where $x_1$ is the conversion and $x_2$ the temperature, $Da$ the
Damkoehler number, $B$ the heat of reaction, $\beta$ the heat transfer
coefficient, $\gamma$ the activation energy, and $x_{2c}$ the coolant
temperature. The dimension-less Damkoehler number $Da$ plays an
important role in determining the qualitative system behavior and will
be assumed to be a hidden parameter. All other parameters are fixed:
$B=22.0$,
$\beta=3.0$, $\gamma=12.0$, and $x_{2c}=0.5$.

We restrict the range of the hidden $Da$ number to be within $\pm 10\%$ of the value
$0.078$. This is an intentional choice, as $Da=0.078$ is the critical
value at which the system exhibits bifurcation behavior: the system
reaches steady state when $Da<0.078$ and limit cycle state when
$Da>0.078$.

To generate the training data set, we set the domain-of-interest for
the state variables to be $(x_1, x_2)\in
[0.1,1.0]\times[0.5,5.5]$. The time step is set as $\Delta
t=0.02$. Upon conducting numerical tests, we set the memory step to
$n_M=700$ and the recurrent step to $n_R=1$.

\begin{figure}
	\centering
	\begin{subfigure}[b]{0.5\textwidth}
		\includegraphics[width=\textwidth]{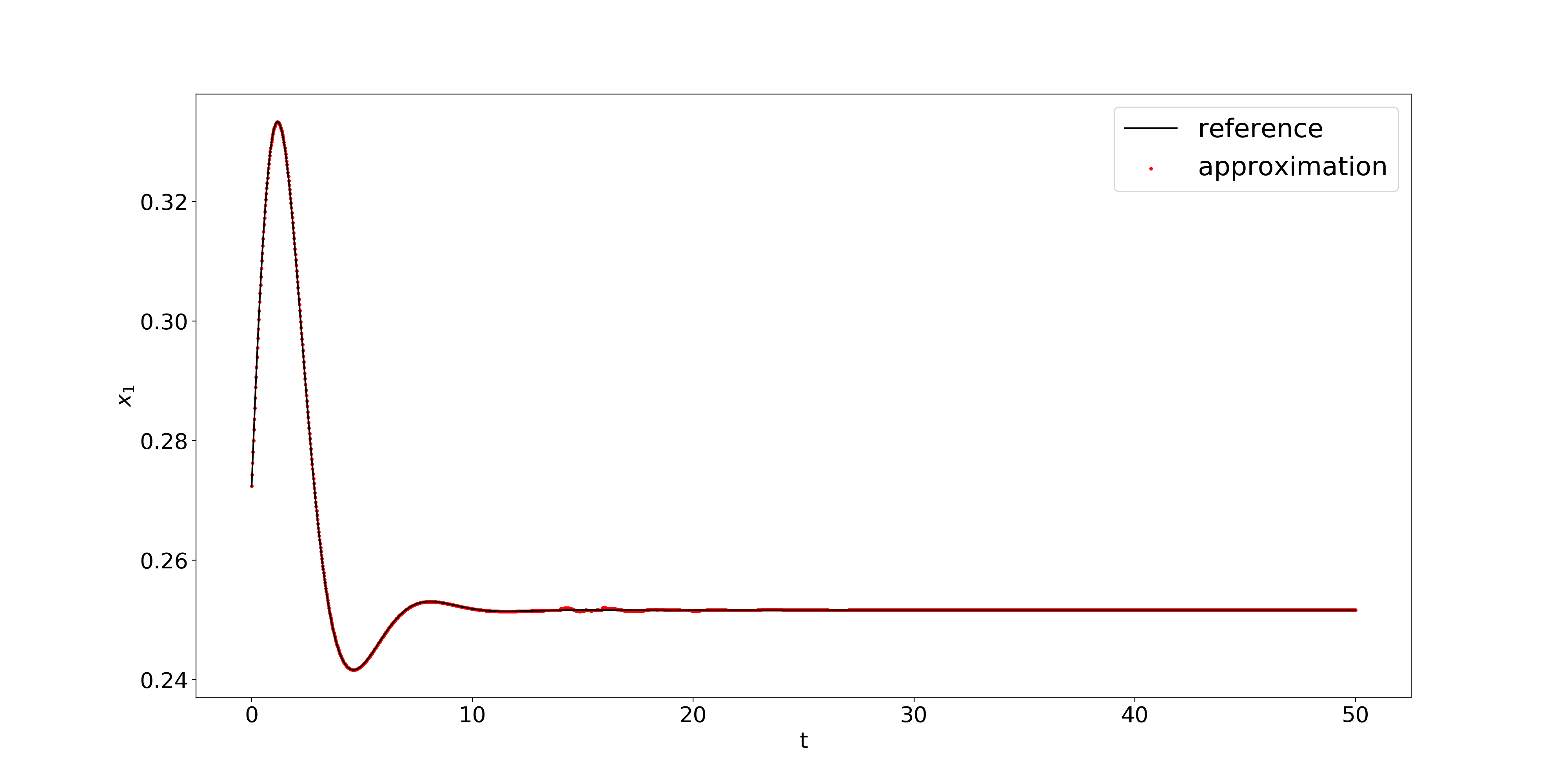}
		\caption{Case 1: $x_1$}
	\end{subfigure}%
	\begin{subfigure}[b]{0.5\textwidth}
		\includegraphics[width=\textwidth]{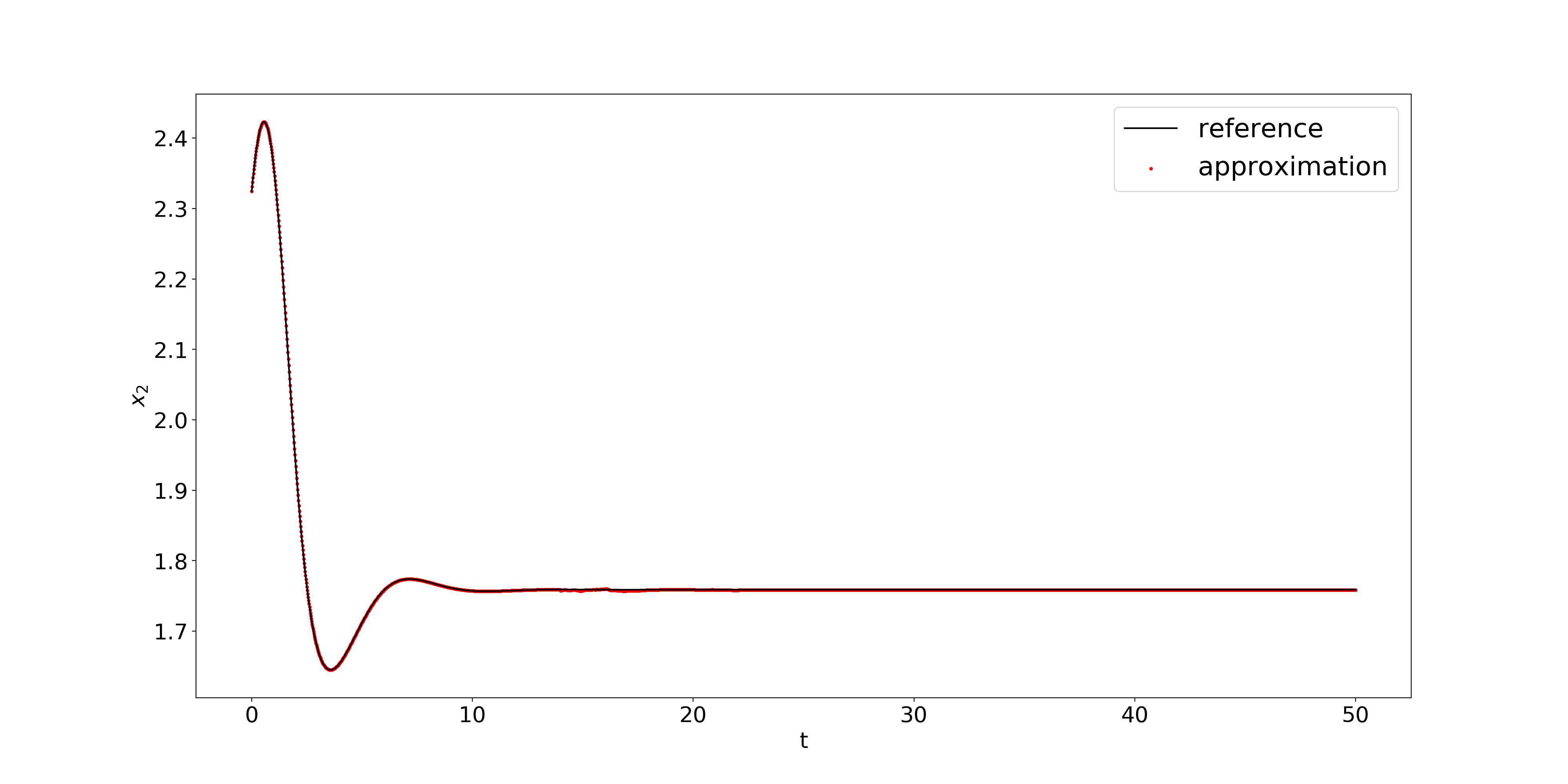}
		\caption{Case 1: $x_2$}
	\end{subfigure}%
	\hfill
	\begin{subfigure}[b]{0.5\textwidth}
		\includegraphics[width=\textwidth]{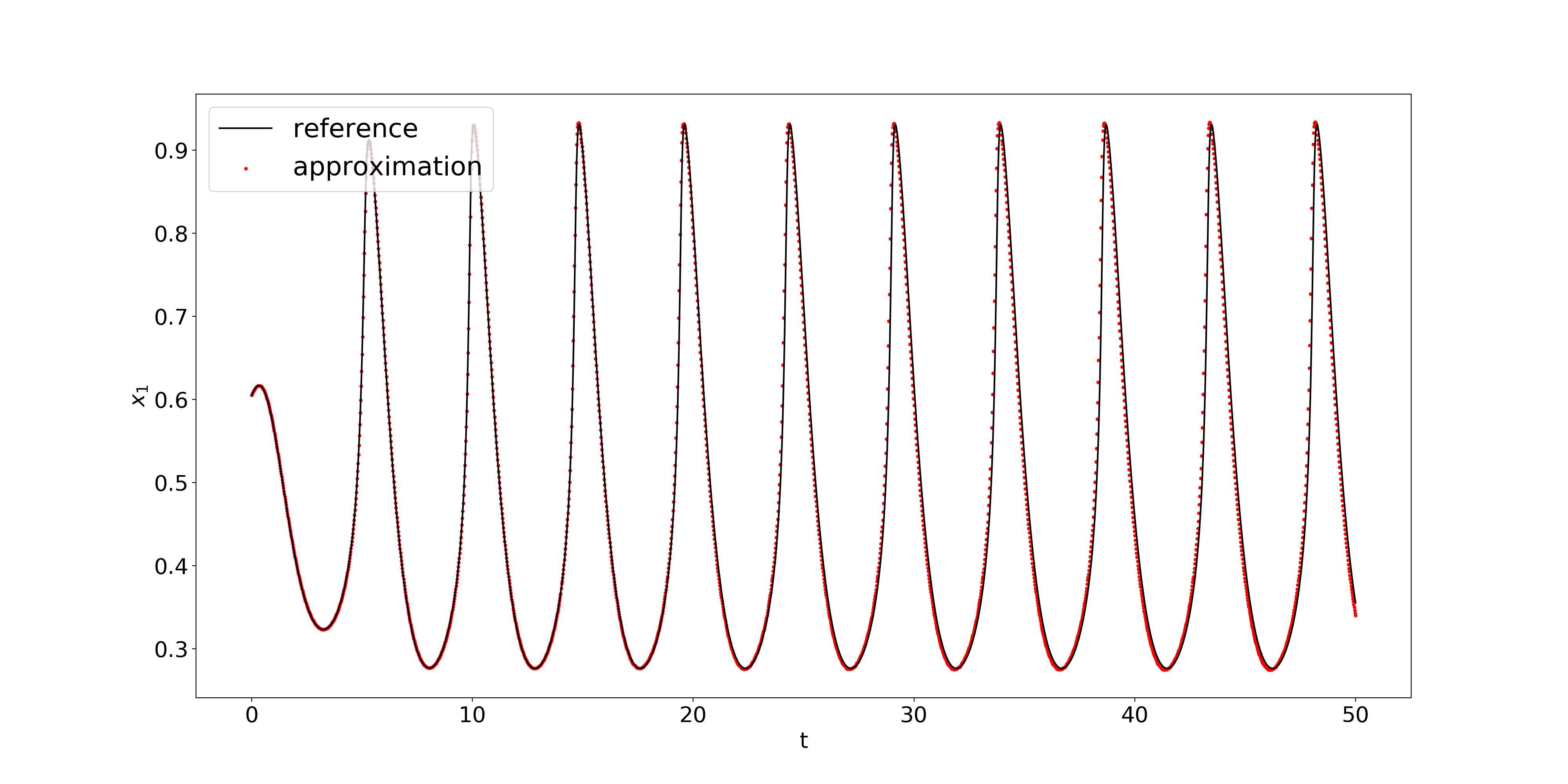}
		\caption{Case 2: $x_1$}
	\end{subfigure}%
	\begin{subfigure}[b]{0.5\textwidth}
		\includegraphics[width=\textwidth]{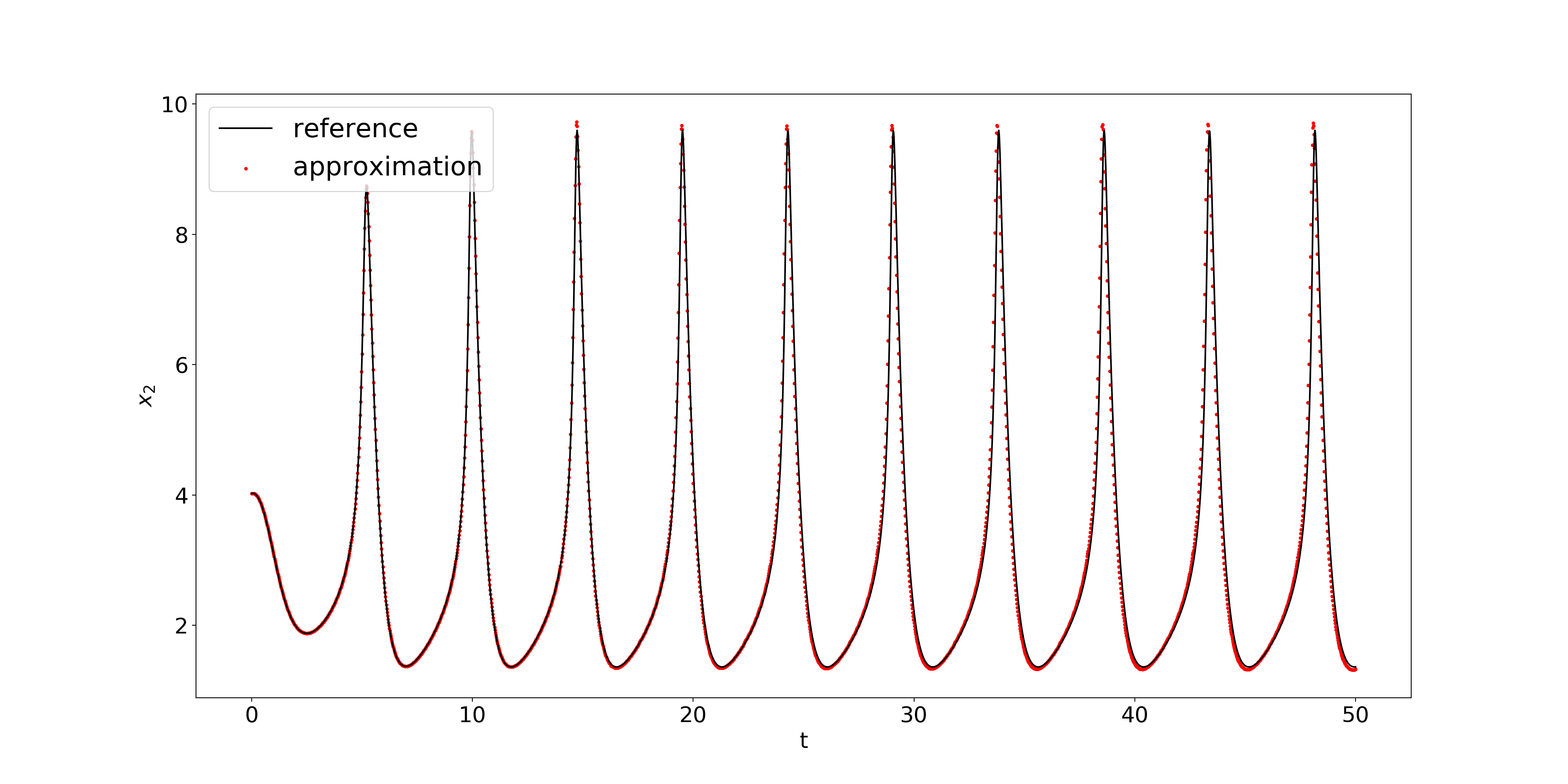}
		\caption{Case 2: $x_2$}
	\end{subfigure}%
	\hfill
	\begin{subfigure}[b]{0.5\textwidth}
		\includegraphics[width=\textwidth]{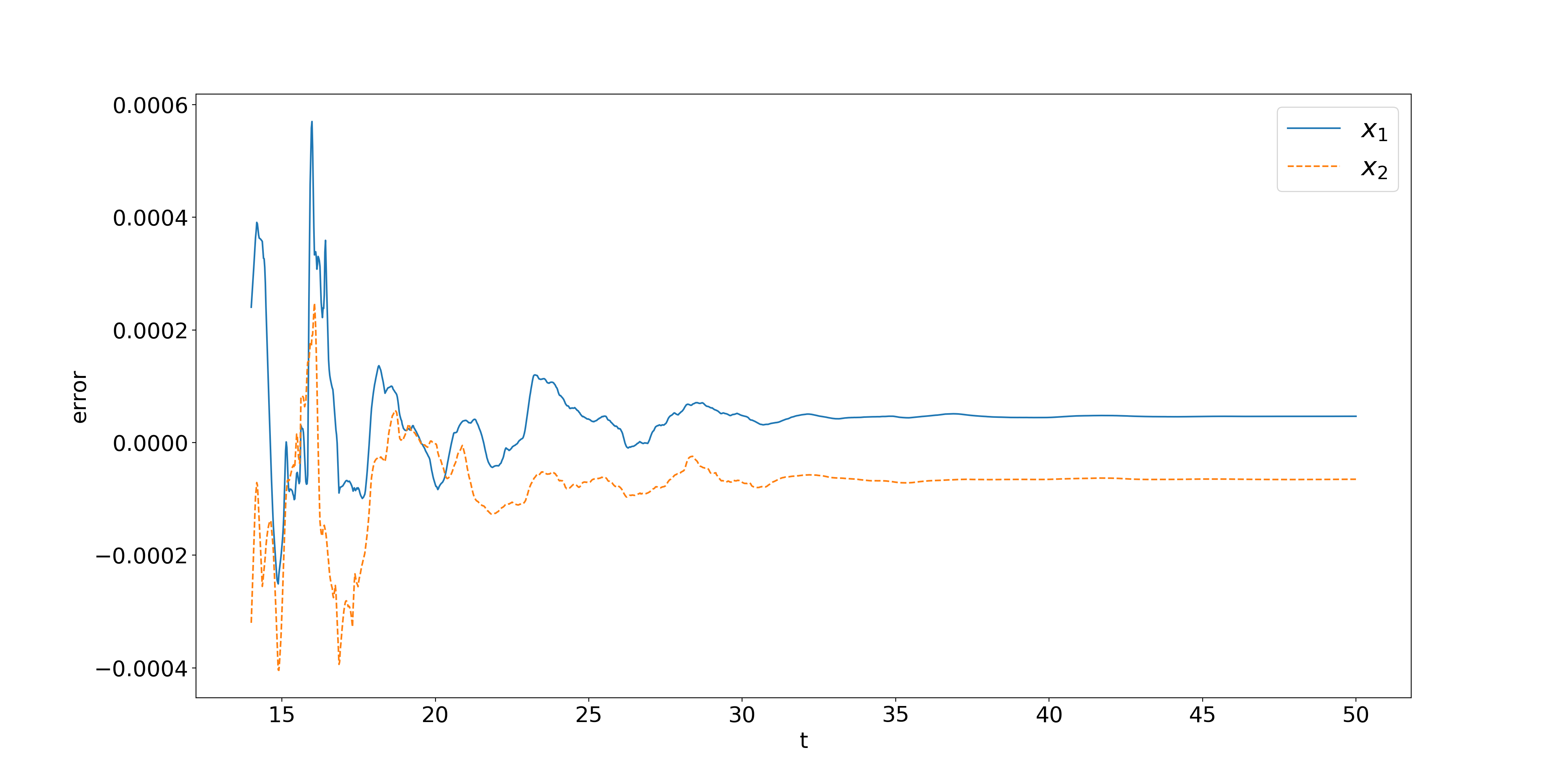}
		\caption{Case 1: relative error.}
	\end{subfigure}%
	\begin{subfigure}[b]{0.5\textwidth}
		\includegraphics[width=\textwidth]{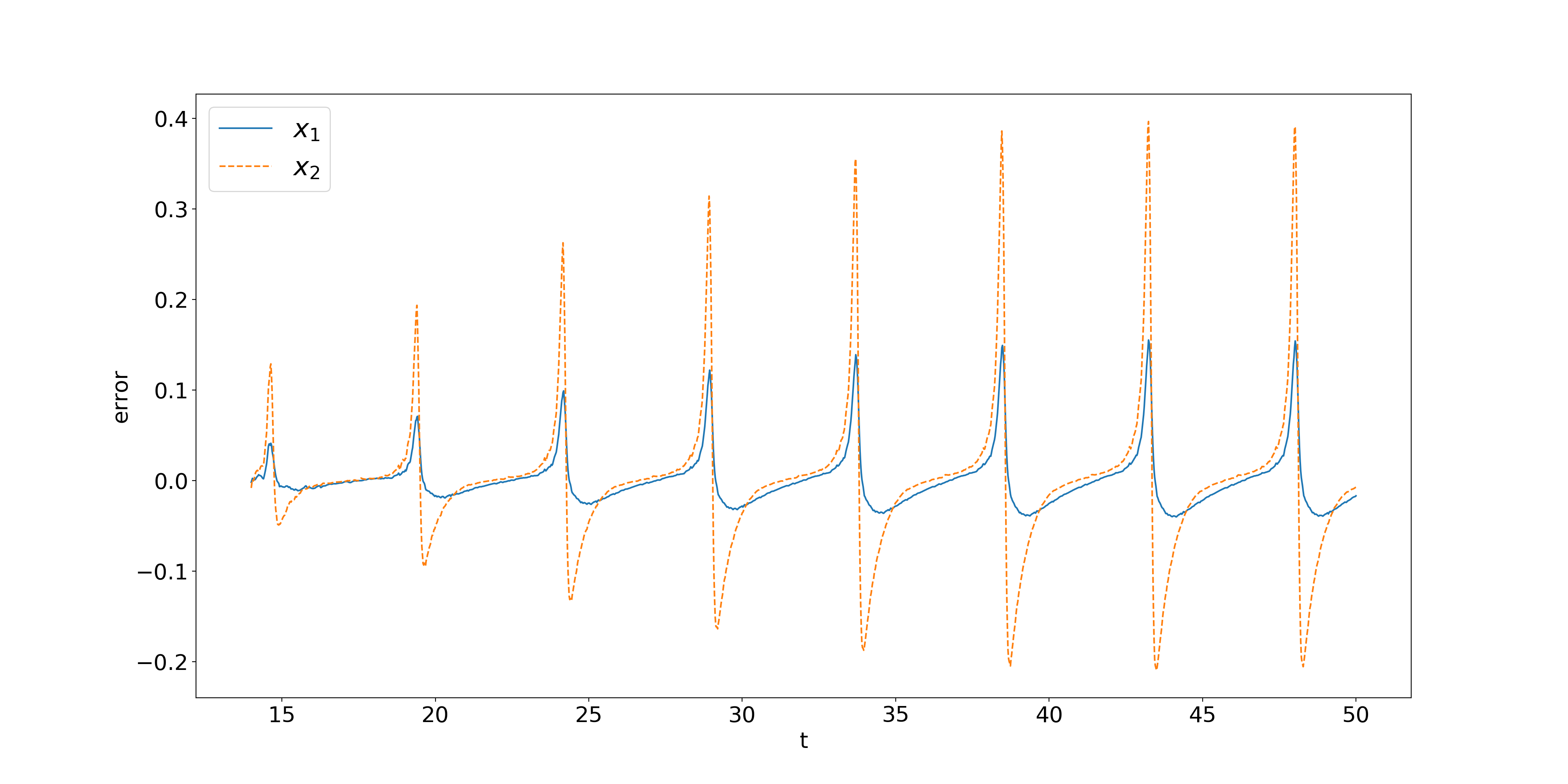}
		\caption{Case 2: relative error.}
	\end{subfigure}%
	\caption{Example 3. Model predictions up to $t = 50$ with
          $n_M=700$ and $n_R=1$ with two cases of arbitrarily chosen
          initial conditions and system parameters.}
	\label{fig:CSTR_preds}
\end{figure}

\begin{figure}
	\centering
	\begin{subfigure}[b]{0.7\textwidth}
		\includegraphics[width=\textwidth]{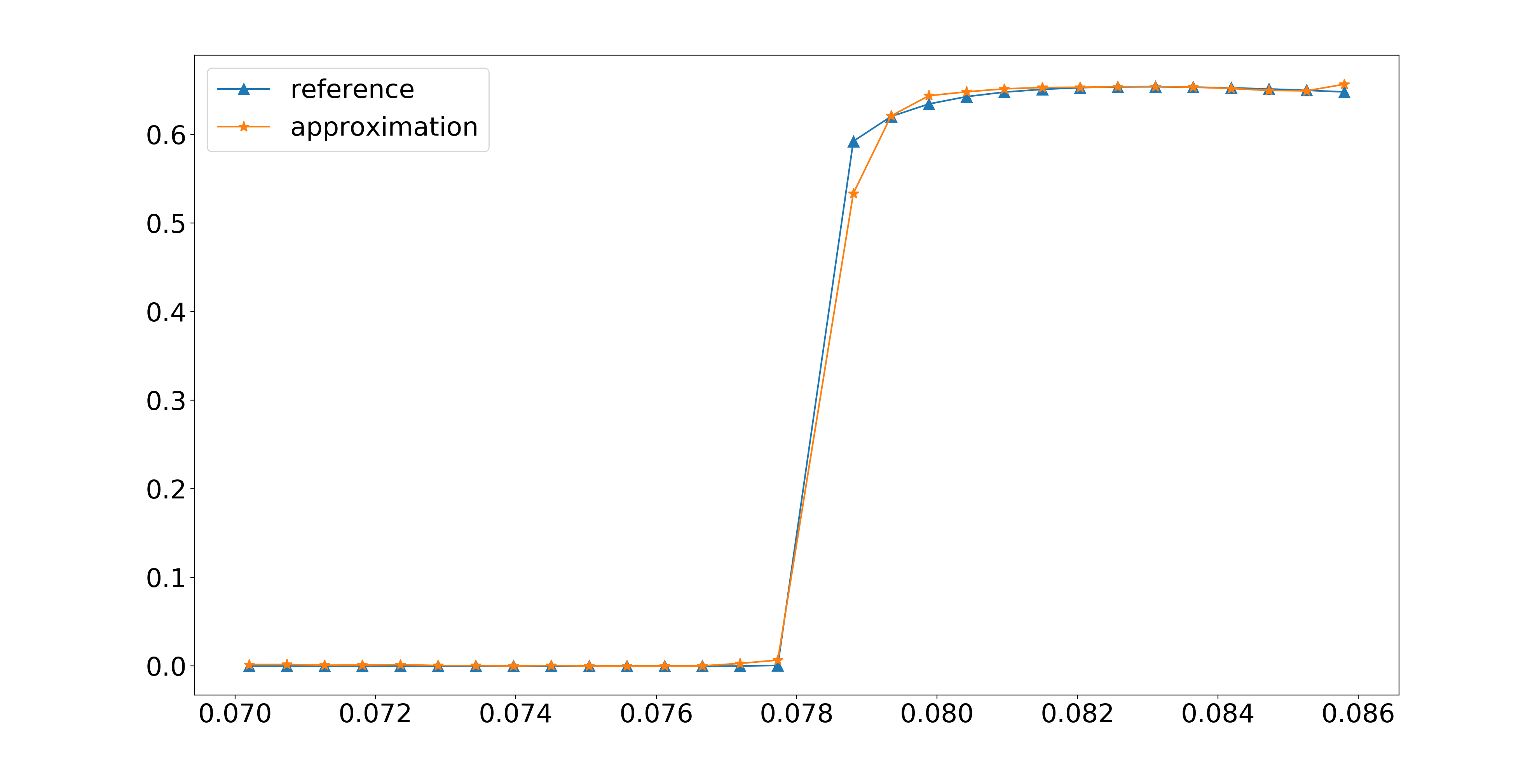}
		\caption{Amplitude of $x_1$ {\em vs.} $Da$.}
	\end{subfigure}
	\hfill
	\begin{subfigure}[b]{0.7\textwidth}
		\includegraphics[width=\textwidth]{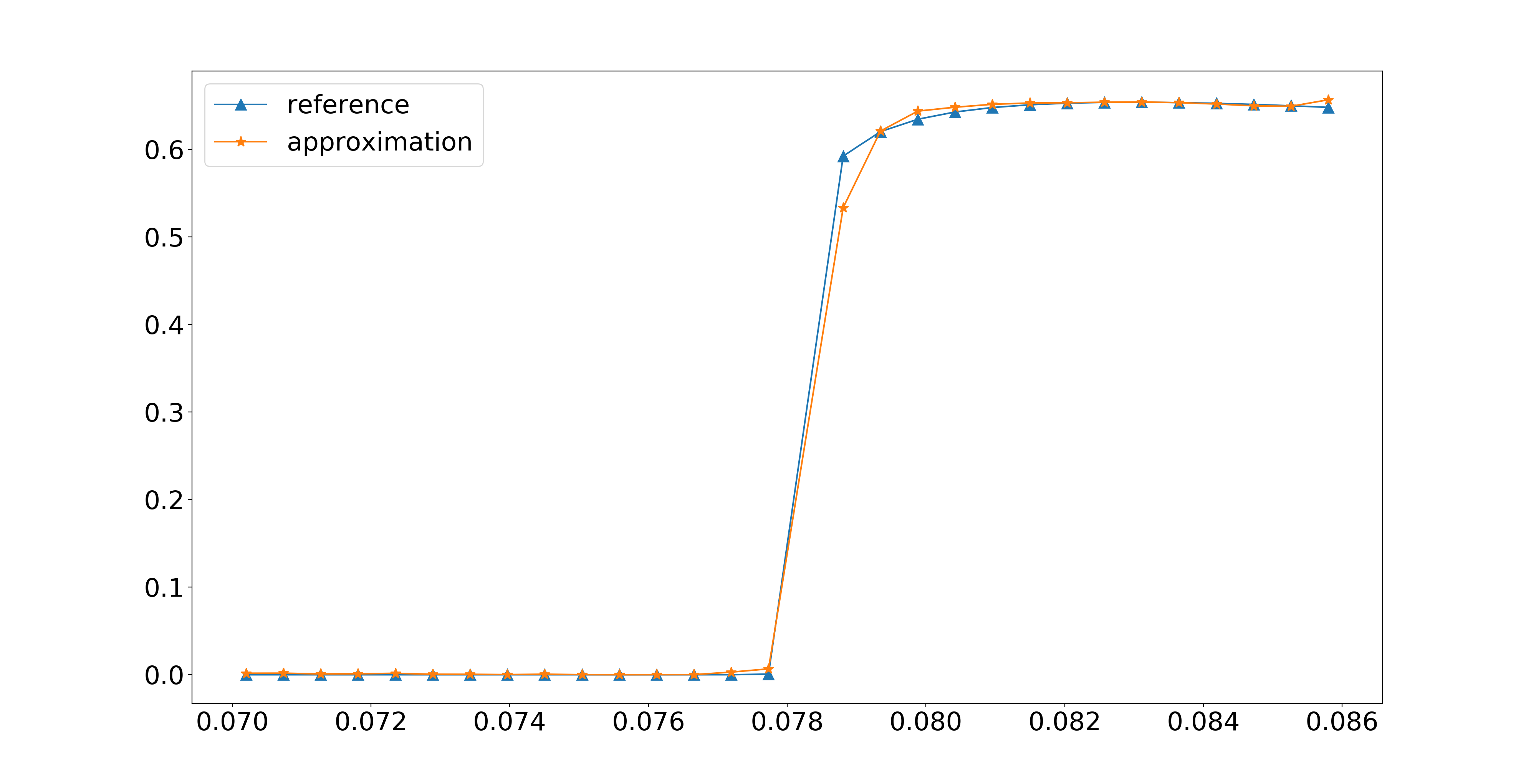}
		\caption{Amplitude of $x_2$ {\em vs.} $Da$.}
	\end{subfigure}
	\caption{Example 3. Solution amplitudes at limiting states
          with respect to $Da$ number.}
	\label{fig:bifurcation}
\end{figure}

We show the DNN trajectory predictions in Fig.~\ref{fig:CSTR_preds},
with two sets of arbitrarily chosen initial conditions and parameters
where trajectories exhibiting steady state and limit cycle
respectively. We observe the predictions match the reference solutions
very well in both cases.
To determine the qualitative behavior of the solutions, we compute the
amplitude of the solutions when they reach a stable state over
a relatively longer time interval $t\in[50,70]$. If the
trajectory reaches a steady state, then the amplitude approaches $0$; if the trajectory
becomes periodic, then its amplitude approaches a constant value.
Fig.~\ref{fig:bifurcation} shows the amplitudes of the 
predictions with respect to the value of $Da$, for both $x_1$ and
$x_2$. We clearly observe the transition from steady state to periodic
state when $Da\approx 0.078$. The comparison between the DNN
predictions and the reference true solutions again shows good
agreement.

\subsection{Example 4: Cell signaling cascade}

We consider a dynamical system model for autocrine cell-signaling
loop. The 3-dimensional state variable $[\ea, \eb,\ec]^\top$ denotes
the dimensionless concentrations of the active form of the
enzymes. The true (and unknown) governing equations are 
\be \label{govern_cell}
\left\{
\begin{split}
	&\frac{d\ea}{dt} = \frac{I(t)}{1+G_4\ec}\frac{V_{max,1}(1-\ea)}{K_{m,1}+(1-\ea)} - \frac{V_{max,2}\ea}{K_{m,2}+\ea}, \\
	&\frac{d\eb}{dt} = \frac{V_{max,3}\ea(1-\eb)}{K_{m,3}+(1-\eb)} - \frac{V_{max,4}\eb}{K_{m,4}+\eb},\\
	&\frac{d\ec}{dt} = \frac{V_{max,5}\eb(1-\ec)}{K_{m,5}+(1-\ec)} - \frac{V_{max,6}\ec}{K_{m,6}+\ec},\\
\end{split}
\right.
\ee
where $I=1.0$, $G_4=0.2$ are fixed and the parameters $K_{m,i}$, $V_{max,i}$,
$i=1,\dots,6$, are hidden parameters, for a total of
12 hidden parameters. For this study, we restrict the hidden parameters
to within $\pm 10\%$ of their nominal values. The nominal values for
all $K_{m,i}$, $i=1,\dots, 6$, are fixed at 0.2, and for $V_{max,1}$
is 0.5, for $V_{max,2,3,4}$ are 0.15, for $V_{max,5}$ is 0.25, and
for $V_{max,6}$ is 0.05.
The domain-of-interest for the state variable is
$[0,1]^3$. 

\begin{figure}
	\centering
	\begin{subfigure}[b]{0.5\textwidth}
		\includegraphics[width=\textwidth]{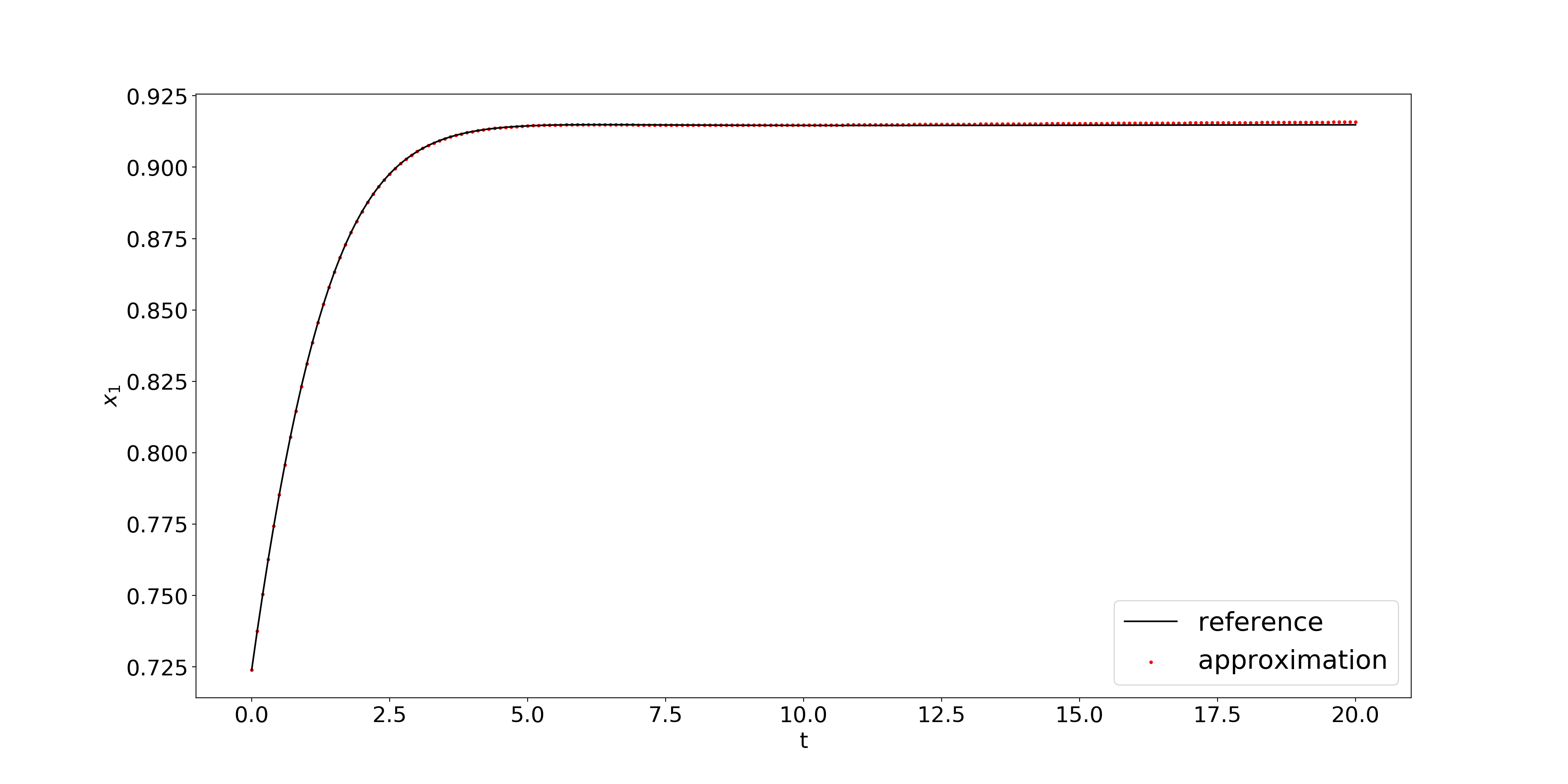}
		\caption{$x_1$}
	\end{subfigure}%
	\begin{subfigure}[b]{0.5\textwidth}
		\includegraphics[width=\textwidth]{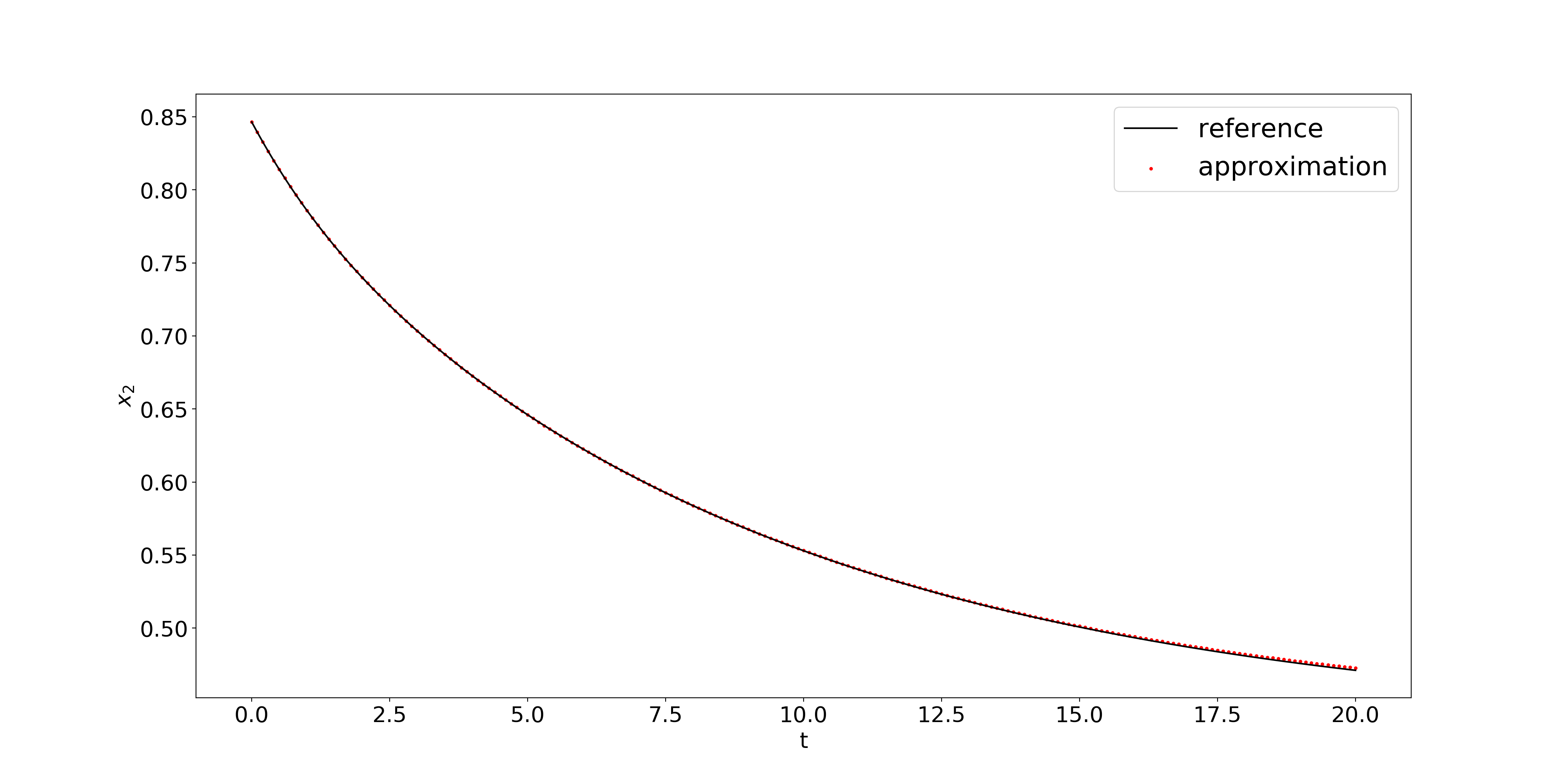}
		\caption{$x_2$}
	\end{subfigure}%
	\hfill
	\begin{subfigure}[b]{0.5\textwidth}
		\includegraphics[width=\textwidth]{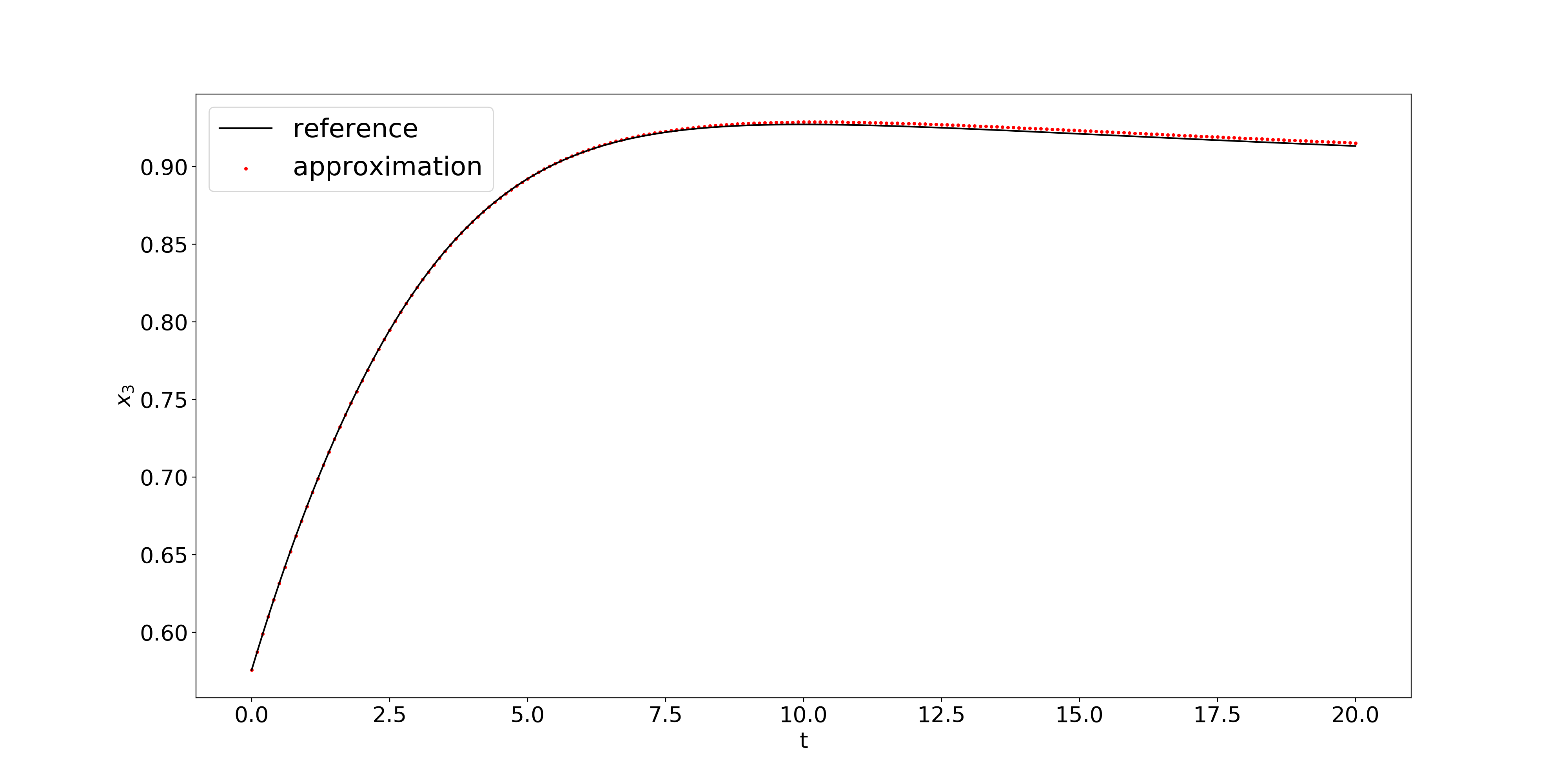}
		\caption{$x_3$}
	\end{subfigure}%
	\begin{subfigure}[b]{0.5\textwidth}
		\includegraphics[width=\textwidth]{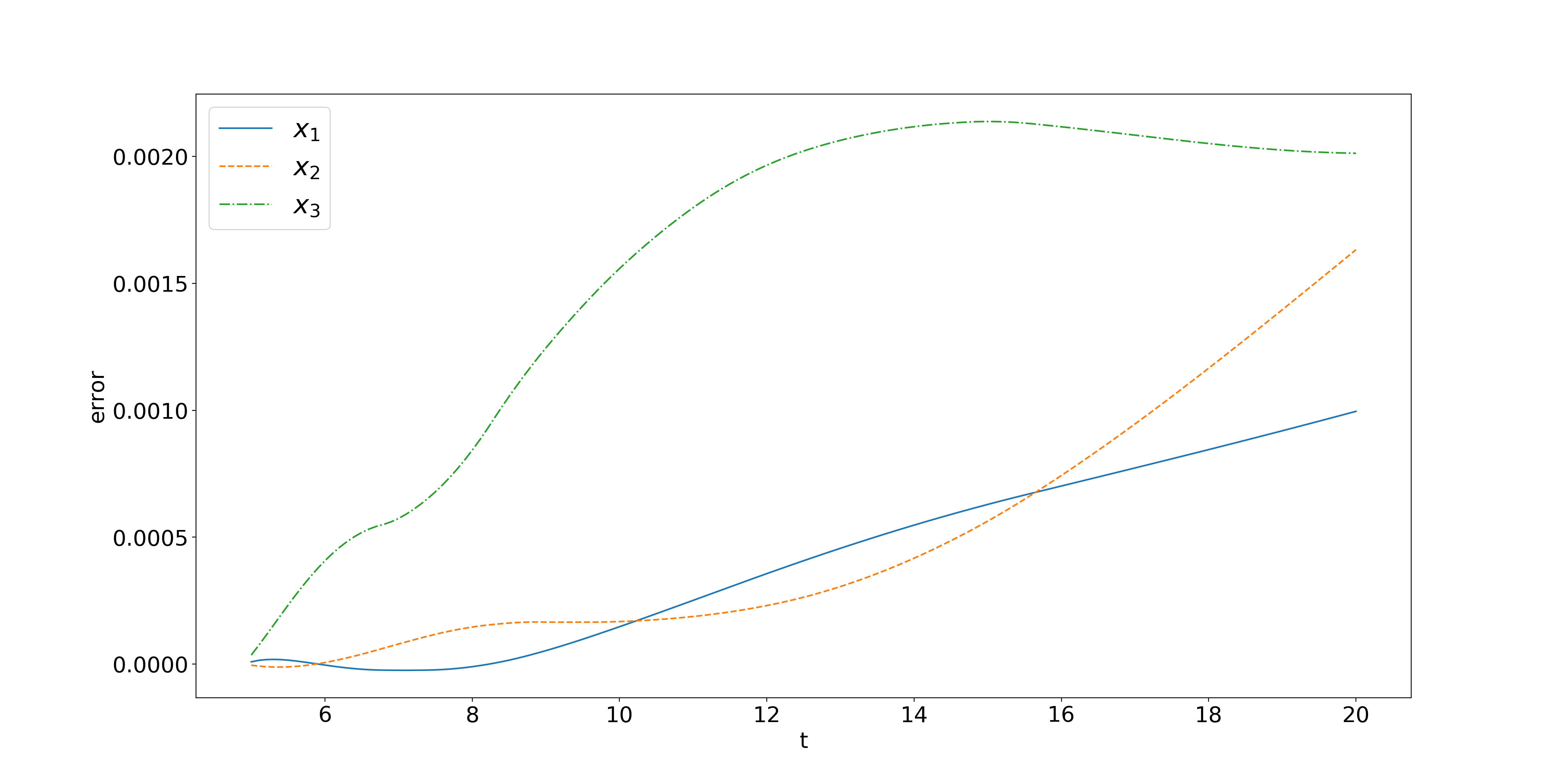}
		\caption{error}
	\end{subfigure}%
	\hfill
	\caption{Example 4. Model predictions and errors up to $t=20$ with $n_M=50$ and $n_R=12$ using a set of arbitrary initial conditions and parameters.}
	\label{fig:cell_preds}
\end{figure}

The training data are constructed by collecting $2$ randomly selected
sequences of consecutive data entries from $75,000$ trajectories,
generated by uniformly distributed random initial conditions over
$300$ steps with a time step $\Delta t = 0.1$. In our DNN model, the
memory steps is set as $n_M=50$ and the recurrent steps as $n_R=12$.
The trajectory predictions and the error
plots are shown in Fig.~\ref{fig:cell_preds}, with a set of arbitrary
initial conditions and system parameters. We observe that the DNN
predictions match the reference solutions very well for
up to $t=20$. 

\section{Conclusion} \label{sec:conclusions}

We presented a deep learning strategy for modeling unknown dynamical
systems with hidden parameters. By incorporating both memory terms in
the network input layer and
recurrent terms in the network loss function computation, the proposed
DNN is able to learn the unknown flow map of the system, by only using
trajectory data of the state variables. A distinct feature of the DNN
structure is that it is able to model the system with completely
hidden and unknown parameters. This can be useful for practical
problems, where many system parameters can not be measured. The
proposed DNN method thus provides a highly flexible approach for learning
unknown dynamical systems.

\bibliographystyle{siamplain}
\bibliography{LearningEqs,neural}

\appendix

\section{Details of Example 2 in Section \ref{sec:example2}}

The detailed setting of Example 2 is
$\x=(\p; \q)$, where $\p\in\Rs^{10}$ and $\q\in\Rs^{10}$ satisfy
\begin{equation}
\begin{cases}
\dot \p = \Sigma_{11} \p + (\I + \Sigma_{12}) \q,\\
\dot \q = -(\I + \Sigma_{21}) \p - \Sigma_{22} \q.
\end{cases}       
\end{equation}
Here, $\I$ is the identity matrix of size $10\times
10$, and $\Sigma_{ij} \in \Rs^{10\times 10}$,
$i=1,2,j=1,2$ are four coefficient matrices. We set three of the
coefficient matrices as fixed, with $\Sigma_{11} = \Sigma_{12} = \mathbf{0}$, and
\begin{frame}
	\footnotesize
	\setlength{\arraycolsep}{4.4pt} 
	\begin{align*} 
		&\Sigma_{22}  \times 10^3 =\\
		&
		\left( \begin{array}{rrrrrrrrrr} 
			1500 & 124 & 814 & -104 & -179 & -223 & -731 & -189 & -400 & 242\\
			124 & 836 & 679 & 277 & 197 & -515 & -52.1 & -273 & 101 & 301\\
			814 & 679 & 1500 & 651 & 755 & -605 & -379 & -546 & -225 & 223\\
			-104 & 277 & 651 & 1960 & 720 & -782 & -299 & -775 & -180 & 506\\
			-179 & 197 & 755 & 720 & 2290 & -973 & 518 & -19.1 & -604 & -369\\
			-223 & -515 & -605 & -782 & -973 & 1290 & -400 & 412 & 314 & -420\\
			-731 & -52.1 & -379 & -299 & 518 & -400 & 1960 & 68.3 & 455 & -316\\
			-189 & -273 & -546 & -775 & -19.1 & 412 & 68.3 & 576 & -53.6 & -332\\
			-400 & 101 & -225 & -180 & -604 & 314 & 455 & -53.6 & 1030 & 265\\
			242 & 301 & 223 & 506 & -369 & -420 & -316 & -332 & 265 & 1090
		\end{array} \right).
	\end{align*}
      \end{frame}
      The 100 entries of the matrix $\Sigma_{21}$ 
are treated as hidden parameters.

\end{document}